\documentclass[conference,10pt]{IEEEtran}

\AtBeginDocument{%
  \providecommand\BibTeX{{%
    \normalfont B\kern-0.5em{\scshape i\kern-0.25em b}\kern-0.8em\TeX}}}

 \usepackage{pifont}
\usepackage{tabularx}
\usepackage{booktabs} 
\usepackage{adjustbox}
\usepackage{amsmath}

\usepackage{url}
\usepackage{enumitem}

\usepackage[utf8]{inputenc} 
\usepackage[T1]{fontenc}    
\usepackage{hyperref}       
\usepackage{url}            
\usepackage{booktabs}       
\usepackage{amsfonts}       
\usepackage{nicefrac}       
\usepackage{microtype}      
\usepackage{xcolor}         
\usepackage{colortbl}
\usepackage{subcaption}
\usepackage{graphicx}
\usepackage{svg}
\usepackage{multirow}
\usepackage{wrapfig}
\usepackage{threeparttable}
\usepackage{tikz}
\usepackage{makecell}
\usepackage{tablefootnote}
\usepackage{cite}
\usepackage{color}
\usepackage{amsmath}
\usepackage{float}

\begin{document}
\title{Accelerating Exact Combinatorial Optimization via RL-based Initialization -- A Case Study in Scheduling}

\newcommand{\red}[1]{\textcolor{red}{#1}}

\author{\IEEEauthorblockN{Jiaqi Yin}
\IEEEauthorblockA{
\textit{University of Utah}\\
Salt Lake City, US \\
jiaqi.yin@utah.edu}
\and
\IEEEauthorblockN{Cunxi Yu}
\IEEEauthorblockA{
\textit{University of Utah}\\
Salt Lake City, US \\
cunxi.yu@utah.edu}
}

\maketitle

\begin{abstract}

Scheduling on dataflow graphs (also known as computation graphs) is an NP-hard problem. The traditional exact methods are limited by runtime complexity, while reinforcement learning (RL) and heuristic-based approaches struggle with determinism and solution quality. This research aims to develop an innovative approach that employs machine learning (ML) for addressing combinatorial optimization problems, using scheduling as a case study. The goal is to provide guarantees in optimality and determinism while maintaining the runtime cost of heuristic methods. Specifically, we introduce a novel two-phase RL-to-ILP scheduling framework, which includes three steps: 1) RL solver acts as coarse-grain scheduler, 2) solution relaxation and 3) exact solving via ILP. Our framework demonstrates the same scheduling performance compared with using exact scheduling methods while achieving up to 128 $\times$ speed improvements. This was conducted on actual EdgeTPU platforms, utilizing ImageNet DNN computation graphs as input. Additionally, the framework offers improved on-chip inference runtime and acceleration compared to the commercially available EdgeTPU compiler.

\end{abstract}

\maketitle

\section{Introduction}


Combinatorial optimization (CO) is an optimization problem that finds the optimal solution from a vast search space, where graph-based scheduling is one of the classic CO problems: allocate the operators in computational graph into the given number of stages and minimize the on-cache memory, and communication cost simultaneously.
Computational graph scheduling is fundamental to a wide range of hardware deployment domains such as FPGAs, CPU/GPU, and domain-specific accelerators. For example, the DNN model execution is computationally intensive, which requires many computational resources including on-cache memory, I/O-bandwidth, etc \cite{jouppi2017datacenter}. This problem becomes more critical for edge devices considering the small buffer size and fewer processing elements. More importantly, in the increasing uses of cloud-based domain-specific accelerators, enabling a fast runtime compilation (scheduling) is critical to the global management of cloud computing resources in efficiency and cloud utilization \cite{rittinghouse2017cloud}. All together make scheduling an important task in pursuing both optimal quality of results and fast runtime. 

\textbf{Challenges of graph-level scheduling:} 
To clearly motivate this work, we provide a simple case study by evaluating conventional scheduling methods, including exact methods \cite{leiserson1991retiming,zhang2013sdc}, heuristic methods (commercial EdgeTPU compiler), and stand-alone RL/ML based scheduling methods. Specifically, we pick Google Edge TPU as the case study platform, with ImageNet model ResNet152 \cite{he2016deep}. The properties of the three scheduling methods are presented in Table \ref{tbl:methodologies_Comparison}.
The key conclusions are summarized as follows: \textbf{1)} The advantage of exact methods \cite{yin2022exact,leiserson1991retiming,zhang2013sdc} (e.g., ILP or SAT) is the optimality of scheduling results. ILP constructs the formulations and constraints to solve the scheduling problem, which guarantees solution optimality. However, due to the nature of the complexity, exact methods suffer from scalability issues with the highest solving runtime; \textbf{2)}. Heuristic methods \cite{yang1993list,ahn2020ordering} have less solving runtime overhead but the scheduling results are sub-optimal. Similar to the EdgeTPU compiler, many vendor-specific libraries such as Nvidia cuBLAS, TVM, TensorFlow-Lite, and EdgeTPU compiler \cite{chen2018tvm,abadi2016tensorflow,sanders2010cuda} are built upon the heuristic approach. \textbf{3)} Machine learning (ML) techniques have been applied to many combinatorial problems such as scheduling and other EDA problems \cite{ren2023machine, huang2021machine,yu2019painting,yu2018developing,wu2023gamora,yu2020flowtune}, which try to solve the given problem at inference runtime to accelerate the solving process, mostly aiming at generating near-optimal scheduling with inference runtime \cite{mao2019learning,chen2019deep,sheng2021deep,yin2023respect,chen2019deep}. However, there are well known limitations in lacking determinism and guarantees for the performance and solving procedure.

\begin{table}[ht]
\caption{Comparison of existing scheduling methodologies with motivating example of ResNet152 computation graph.}
\footnotesize
\label{tbl:methodologies_Comparison}
\begin{tabular}{l|l|l|l}
\toprule
              & Runtime (s) & QoR(MB) & Determinism  \\
\midrule
Exact Methods & Highest ($268 \times$)           & Optimal                                   & \checkmark           \\
\midrule
Heuristic     & High ($66 \times$)     &      Sub-optimal                                       & \checkmark          \\
\midrule
Ours (RL only)   & Lowest ($1 \times$)      & Sub-optimal                                   & \ding{53}          \\
\midrule
Ours (RL+ILP)      & Low ($5.5 \times$)      & Optimal                                   & \checkmark           \\
\toprule
\end{tabular}
\end{table}

\textbf{Our solutions:} In this work, we use DNNs computation graph scheduling as a case study to illustrate the possibility of accelerating exact CO problem solving using a combination of reinforcement learning (RL) and exact solving techniques. Specifically, {we propose Inc-ILP, a novel graph scheduling framework that combines both of merits of RL-based scheduling and the exact method, where RL agent performs coarse-grain solving, and the exact method refines the coarse-grained initial solving and produces the optimal result. Thus, as shown in Table \ref{tbl:methodologies_Comparison}, Inc-ILP generates deterministic and optimal scheduling results with significantly reduced solving runtime cost.} The main contribution of this work can be summarized as follow: \textbf{(1)} we adopt an RL-based framework to imitate the behavior of existing exact methods, which provides initialized scheduling results as output. \textbf{(2)} To refine the RL solution for optimality and determinism, we perform local relaxation w.r.t RL scheduling solution and use exact method to obtain optimal scheduling results. \textbf{(3)} We select Google Edge TPU \cite{boroumand2021mitigating} as the backend and integrate Inc-ILP in an end-to-end compile-to-deploy framework to evaluate real-world performance. \textbf{(4)} Finally, we perform a comprehensive evaluation of Inc-ILP in optimality and solving runtime compared to commercial EdgeTPU compiler (heuristic) and pure exact method \cite{zhang2013sdc}. The results confirm the capability of Inc-ILP in generating optimal and deterministic results with significant speedups. As a result, we believe this work can offer some new intuitions for broadly improving determinism and (semi)optimality for ML in EDA and general combinatorial optimization.

\section{Preliminary}

\begin{figure}[!htb]
    \centering
    \includegraphics[width=0.48\textwidth]{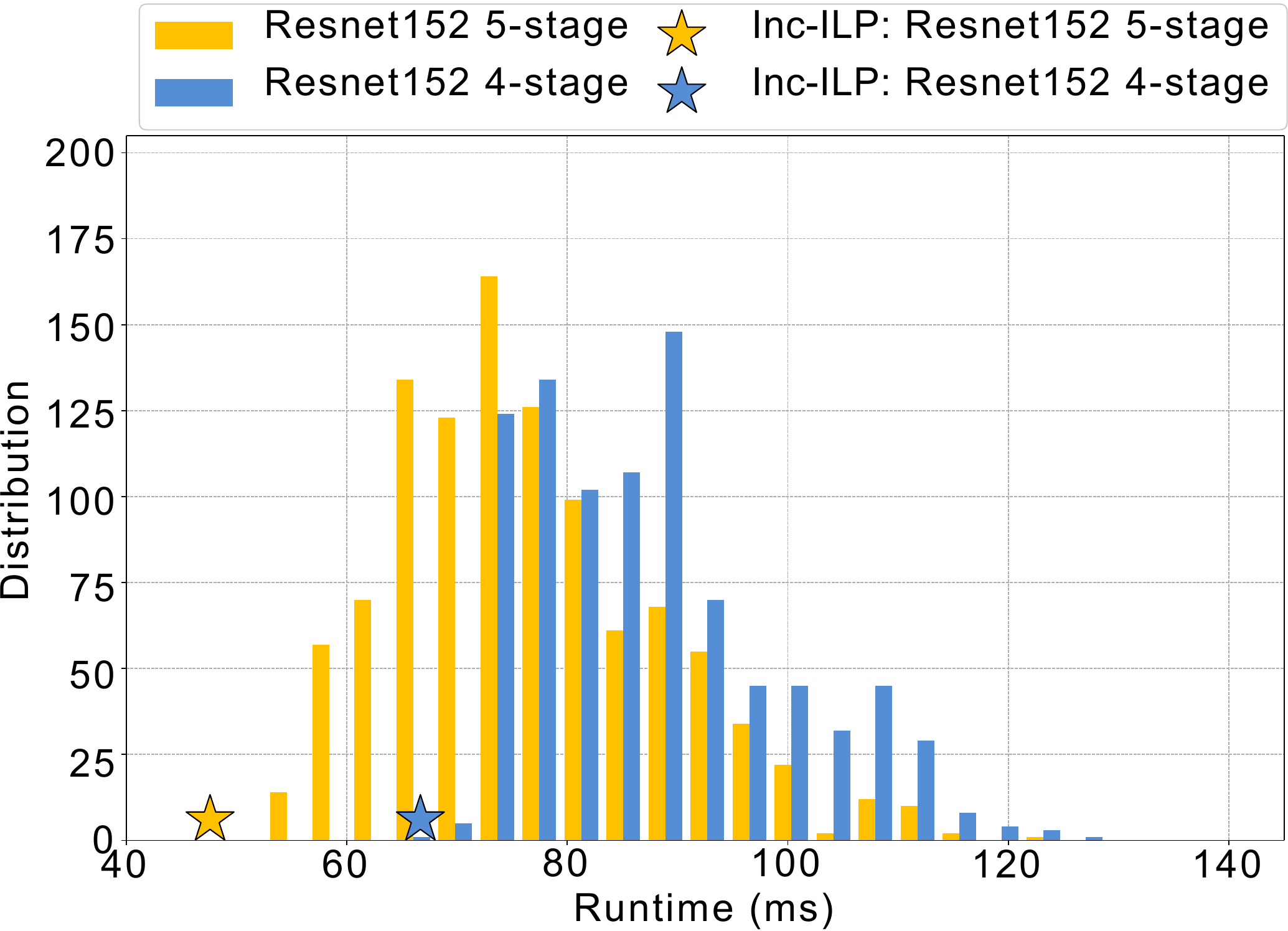}
    \caption{Measurement system illustration of multi-stage pipelined Edge TPUs (physical system setup included in Figure \ref{fig:implementation}). The performance of Inc-ILP scheduling results is located in the bottom-left corner with star markers.}
    \vspace{-3mm}
    \label{fig:motivating_example}
\end{figure}

\subsection{Computational graph scheduling}

\begin{figure}[!htb]
    \centering
    \includegraphics[width=0.45\textwidth]{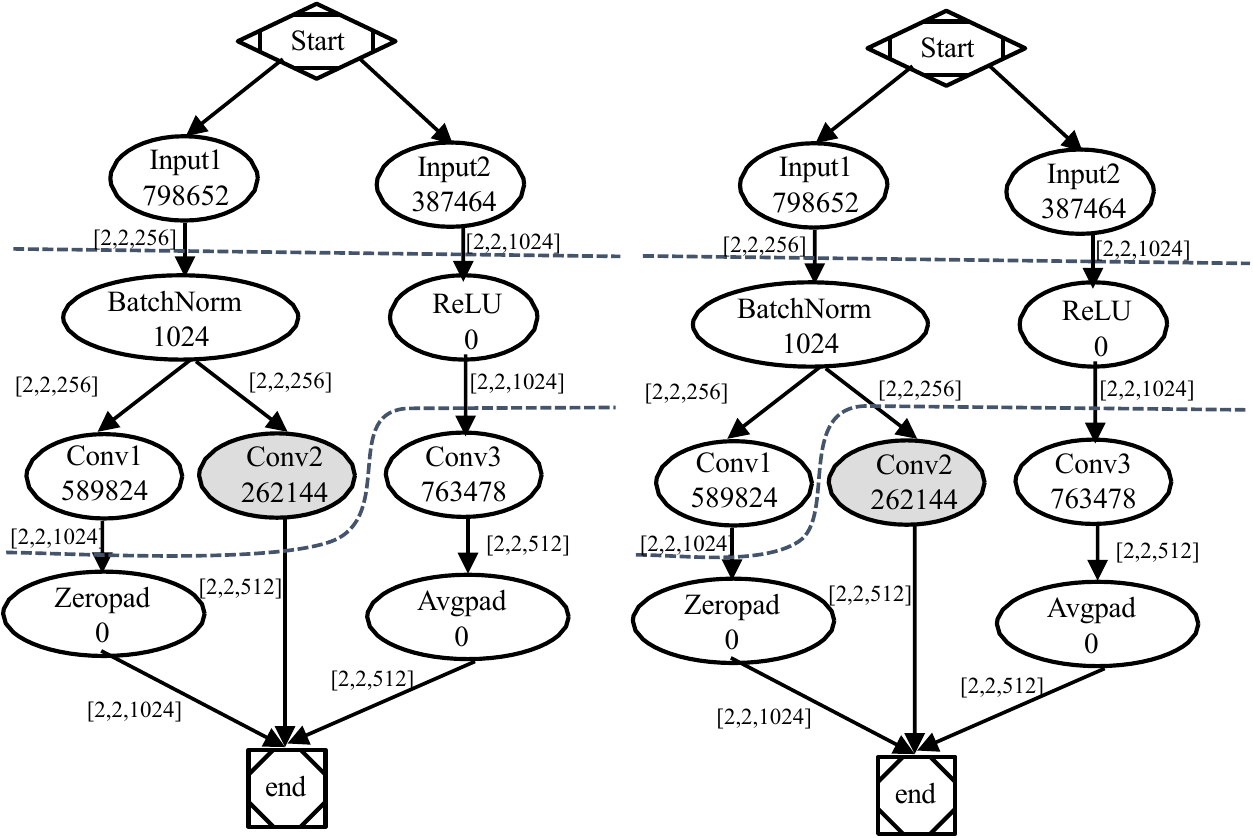}
    \caption{Two distinct scheduling examples for a straightforward DNN computation graph differ solely in the allocation of the Conv2 node - the left scheduling results in reduced off-chip memory overhead, while the right scheduling minimizes device-to-device communication expenses.}
    \label{fig:scheduling_example}
\end{figure}

DNN models can be represented as computation graph $G(V, E)$ with nodes $V$ and edges $E$. The nodes represent the operators in DNN models and the edges describe the dataflow dependency that connects all operators. Specifically, the problem of multi-stage DNN model scheduling can be formulated as follow: given the input of a computation graph and a set of scheduling constraints including the number of pipeline stages and resource constraint (e.g., on-cache memory resource), DNN model scheduling aims to generate the optimal scheduling solution $S$ = {$s_0$,$s_1$,...$s_n$}, where $n \leq |V|$ and $s_n$ represents the stage that node $n$ scheduled. The operators in computation graphs will be deployed to given pipeline stages with the minimum (or maximum) optimization objectives like on-cache memory, communication cost, etc. Figure \ref{fig:scheduling_example} displays a simple 3-stage pipelined Edge TPU scheduling outcome. Considering the left graph, the nodes for \text{start}, \text{Input1} and \text{Input2} are designated to stage-0. Stage-1 is composed of BatchNorm, {ReLU}, {Conv1}, and {Conv2}, while stage-2 incorporates \text{Conv3}, \text{Zeropad}, \text{Avgpad}, and \text{end} nodes. 

While hardware device pipelining can improve DNN execution performance by decreasing the size of off-cache memory, the runtime performance is highly sensitive to scheduling solutions, and finding the optimal one is critical due to limited computation resources including memory, and communication bandwidth. To demonstrate the impacts of scheduling solutions on the pipelining system, we select Google Edge TPU as backends and perform a comprehensive runtime performance analysis using 4-stage, 5-stage ImageNet ResNet152 \cite{he2016deep} models pipelining, which is shown in Figure \ref{fig:motivating_example}. 
We can observe that (1) scheduling solutions make significant differences in pipelined Edge TPU systems, which could result in $2\times$ runtime difference; (2) many scheduling solutions might perform similarly but finding the best performance scheduling is statistically challenging. Hence, there is a great need of developing an near-optimal scheduling approach at low solving complexity. Note that for all two benchmarks, the performance of our proposed Inc-ILP scheduler is located in the left bottom corner which provides near-optimal runtime performance.

{The optimal scheduling solution should aim to minimize costs, such as peak memory usage and communication expenses. Figure \ref{fig:scheduling_example} displays two distinct scheduling approaches for a synthetic DNN computation graph, with the number of parameters represented in the vertices. Although both schedules are valid, they differ in the assignment of the $N_{Conv2}$ node. In the context of a three-stage pipelining system, the total execution time ($T$) of the computational graph is the sum of on-device execution runtime ($T_d$) and communication runtime ($T_c$), such that $T = T_d + T_c$. Here, $T_c$ is primarily influenced by off-chip memory usage, while $T_d$ is affected by the communication expenses, i.e. size of tensors passed between stages. The scheduling solution with lower off-chip memory usage and communication overhead will likely have a faster execution time. It is important to note that in a pipeline scenario, $T_d$ is dominated by the stage with the longest runtime. Among the two scheduling solutions, the left scheduling in Figure \ref{fig:scheduling_example} has a lower memory upper bound (76347 in stage 3), resulting in an improved $T_c$. On the other hand, the right scheduling ([2,2,1024]+[2,2,256]+[2,2,1024]) has a smaller device-to-device communication overhead between stages 2 and 3 compared to the left scheduling solution ([2,2,1024]+[2,2,512]+[2,2,1024]). As a result, the left scheduling has better memory usage and lower communication costs, leading to an optimized $T_d$. In this work, we concentrate on Edge TPU backends where $T_c$ has a greater impact than $T_d$. Our aim is to first optimize the memory upper bound and then minimize device-to-device communication expenses.
}

\subsection{Combinatorial problem solving methods}

Graph scheduling is a NP-hard combinatorial optimization problem and there is no optimal solution in polynomial solving runtime. There existing some traditional approaches including reinforce learning, heuristic, and exact approach, etc.

Reinforcement learning \cite{williams1992simple,bello2016neural} has been applied to many combinatoraial optimization problem \cite{yu2020decision,wu2023gamora,selsam2018learning}. In \cite{bello2016neural}, Bello et al. present an innovative approach to addressing combinatorial optimization challenges through the application of neural networks and reinforcement learning. The authors specifically target the traveling salesman problem (TSP) and develop a recurrent neural network capable of predicting a probability distribution of various city arrangements, given a collection of city coordinates. The method generates high-quality solutions for various combinatorial optimization problems, often surpassing the performance of traditional heuristics and metaheuristics. However, the solution based on refinforcement learning is non-derterministic and lacking optimal performance guarantees.

Several renowned heuristic algorithms exist, such as list scheduling and the force-directed algorithm. The force-directed algorithm incorporates a "force" on each operation and aims to balance computation across all stages by minimizing this "force". List scheduling, in contrast, ranks operations based on specific metrics, scheduling the operation with the highest priority when available. Despite their capabilities, both heuristic algorithms share common limitations: they lack determinism and cannot guarantee the optimal quality of results.

On the other hand, exact approaches \cite{micheli1994synthesis,fan2005cost,ramalingam1999solving,dai2018scalable} include integar linear programming (ILP), satisfiability modulo theories (SMT), and integer difference constraints (SDC). Integer Linear Programming (ILP) and Satisfiability Modulo Theories (SMT) guarantee finding optimal solutions and serve as potent methods for tackling intricate optimization problems. However, they share the same scalability issues as the size and complexity of the problems they address increase. For ILP, the combinatorial nature of integer variables leads to a rapid growth in the number of potential solutions, making it difficult for ILP solvers to find optimal solutions efficiently. The scalability of ILP solvers is further affected by the presence of multiple objectives or constraints, which can significantly increase the search space. On the other hand, SMT solvers deal with the satisfiability of logical formulas with respect to various background theories. As the size and complexity of these formulas grow, the search space for satisfying assignments becomes increasingly large and challenging to navigate. Additionally, SMT solvers may struggle with certain types of problems, such as those involving non-linear arithmetic, which can exacerbate the scalability issue.

\section{Approach}
 
\begin{figure*}[!htb]
    \centering
    \includegraphics[width=1\linewidth]{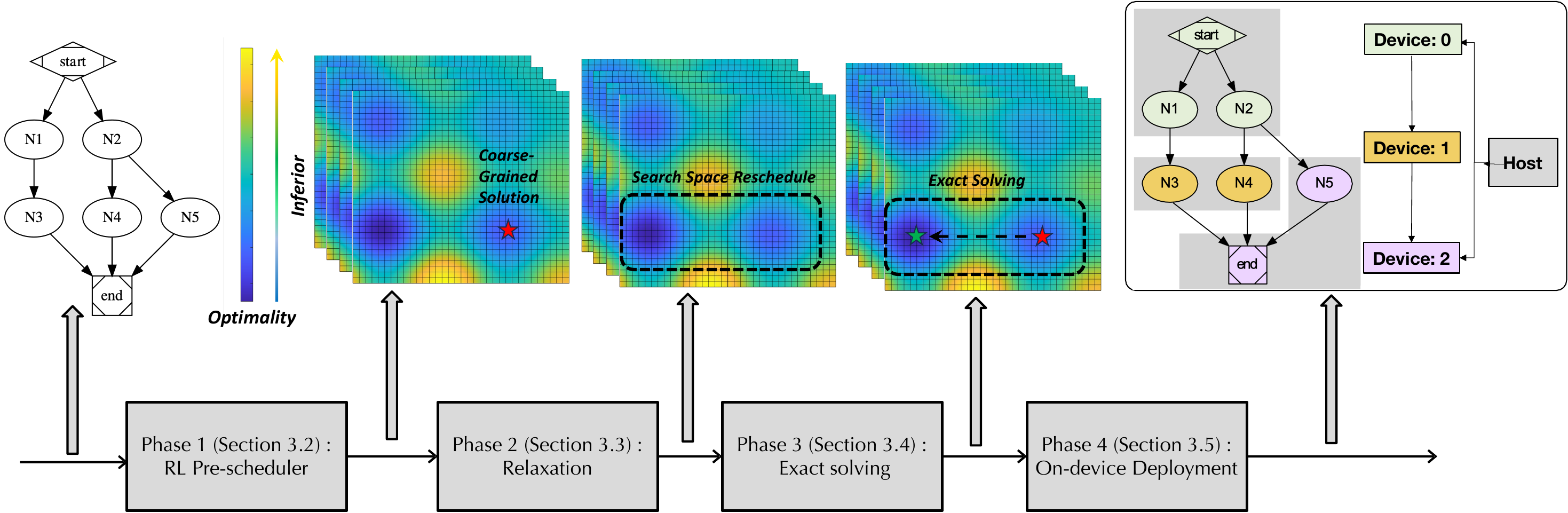}
    \caption{Inc-ILP overview including four-phases: RL pre-scheduling, search space relaxation, exact solving, and device deployment.}
    \label{fig:overflow}
\end{figure*}

\subsection{Overview}

The overall workflow of Inc-ILP including input, output is demonstrated in Figure \ref{fig:overflow}. The Inc-ILP scheduling framework includes four phases: 1) RL pre-scheduling agent generates the coarse-grained scheduling results, 2) search space relaxation based on graph topological order, 3) Inc-ILP produces partially optimal results using exact solving, 4) scheduling results deployment on actual hardware devices.

In phase 1, we adopt the RL model as pre-scheduling agent architecture. We take the benefits of the RL model which can mimic the behavior of exact methods with short inference runtime. Phase 2 performs search space relaxation based on the order of computational graph topological sorting. Then we exactly solve the refined search space and generate the partial optimal solution in phase 3. Specifically, we build ILP constraint formulation to solve the refined search space, which includes dependency constraints, parameter constraints, etc. Finally, in phase 4, we select Google Edge TPU as the backend to deploy scheduling results. The multi-stage pipelined Edge TPU system is shown in Figure \ref{fig:implementation}.

\subsection{Phase 1: RL based pre-scheduling}

 \begin{figure}[!htb]
     \centering
     \includegraphics[width=0.5\textwidth]{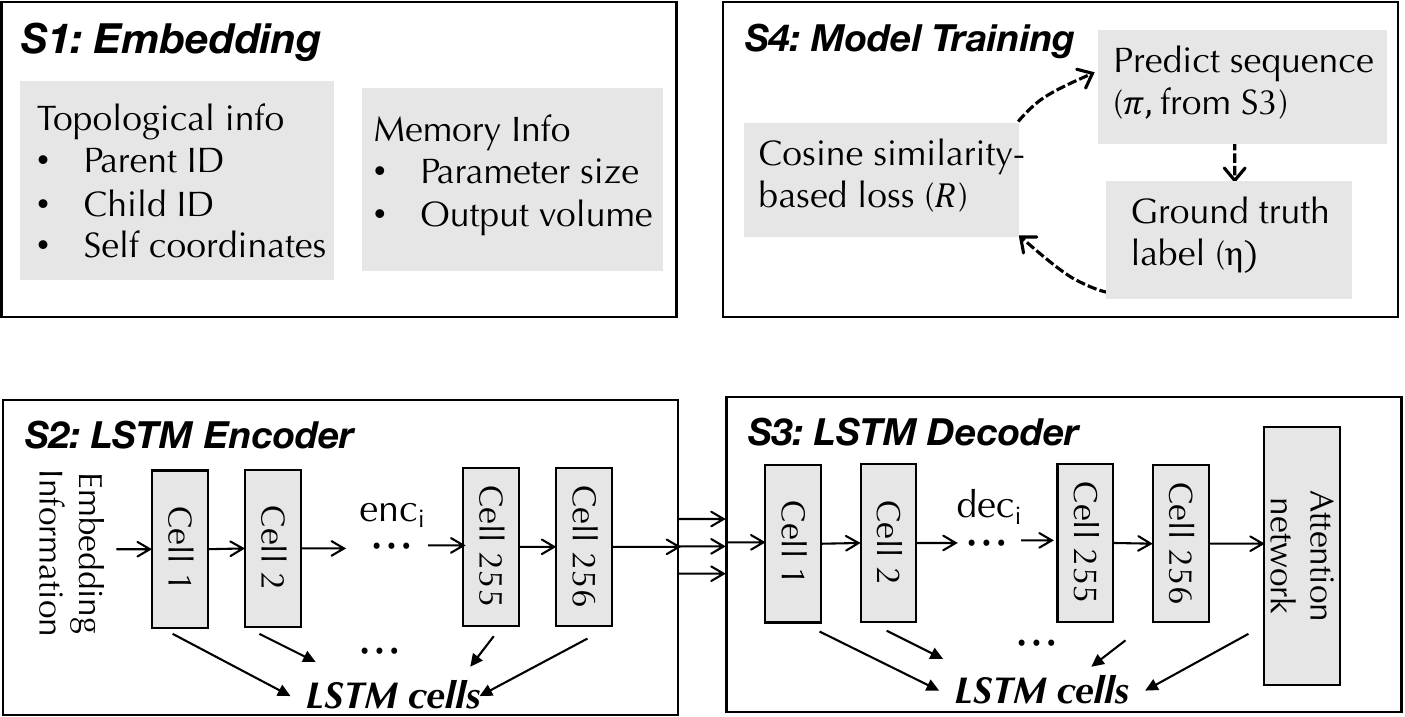}
     \caption{Four-steps Reinforcement Learning (RL) Agent: The encoder processes embedded data, converting it into contextual information, which is then relayed to the decoder. The decoder generates a predicted sequence, and a cosine similarity-based loss is computed to train the RL agent.}
     \label{fig:RL-flow}
 \end{figure}
 
As we mentioned before, RL pre-scheduling generates coarse-grained results, and the quality of results determines if we can achieve global optimality by search space relaxation. Figure \ref{fig:RL-flow} is the expanded illustration of the RL pre-scheduling agent specifications, which are composed of 4 steps (Phase 1.1 - Phase 1.4): embedding, encoder, decoder, and RL model training.

\noindent
\textbf{Computation graph embedding (Phase 1.1) - } The embedding of the computational graph extracts the key information and passes it to the RL model. The graph embedding consists of several components: (1) the absolute and relative coordinates of each node: for node $N_i$, its absolute coordinate is the topological level of $N_i$, and the relative coordinate will be the topological level of parent node $N_i$. Note that we perform depth-first-search (DFS) algorithm to generate topological level, where each node is ordered to the closest position from the source node. For example, the topological order of node $Concat$ in Figure \ref{fig:relaxation} is 2. (2) node IDs: it is generated by hashing all operator names. For node $N_i$, its node ID is $N_i$. (3) memory consumption of each node and its output volume size: scheduling objectives of DNN model pipelining include minimizing peak-memory parameter caching and device-to-device communication. So, memory consumption and output volume size are the most critical attributes of each operator, which can be obtained from model attributes.

\noindent
\textbf{RL agent architecture (Phase 1.2, 1.3) - } As shown in Figure \ref{fig:RL-flow}. The RL agent architecture is composed of an encoder and a decoder. Both the encoder and decoder are built with 256 Long Short-Term Memory (LSTM) cells \cite{hochreiter1997long}. 

\textbf{Encoder - } The encoder digests the graph embedding and transforms it into context information. For each LSTM cell in the encoder, it produces a sequence of latent ($enc_i$) memory states which record the encoding information and propagate along the LSTM dataflow. The last latent memory state will be passed to the decoder as input. 

\textbf{Decoder - } With the computation graph embedding and the latent memory state from the encoder, decoder produces a selection probability distribution over candidate computation nodes using the pointing mechanism. Firstly, similar to the encoder, decoder also generates another latent memory state ($dec_i$) at each LSTM cell. Then the latent memory will be passed to an attention network \cite{vaswani2017attention} to produce candidate node probability distribution. We adopt the pointer network (PtrNet) \cite{bello2016neural} in the attention network, which has been applied to many graph combinatorial optimization problems. Finally, the candidate node probability distribution and latent memory state will be passed to the next LSTM cell. Note that the first decoding latent memory state is a trainable parameter.

\noindent
\textbf{RL model training (Phase 1.4) - } In this step, we iteratively optimize the parameter in the RL agent. We choose the cosine similarity as the basis for constructing the reward function. For each node i in computation graph G(V, E) ($i\le |V|$), we use $\eta(i)$ and $\pi(i)$ to demonstrate the ground truth label and RL agent output sequence respectively. The cosine similarity can be shown in Equation \ref{equ:reward_simi}. Here we introduce a constant $\eta$ to ensure the validity of the equation.

\begin{equation}
\begin{split}
R &= ~\frac{\sum (\pi(i)\cdot \eta(i))}{\max (\sqrt{\sum \pi(i)^2}\cdot \sqrt{\sum \eta(i)^2}, \epsilon)}
\end{split}
\label{equ:reward_simi}
\end{equation}

The reward function for model parameter $\theta$ optimization policy is the maximization of the \textit{cosine similarity}.

\begin{equation}
\label{eq:opt}
\begin{split}
J(\theta | G) &= \mathbb{E}_{\pi\sim p_{\theta}(\cdot | G)}(1-R(\pi | G))
\end{split}
\end{equation}


For ground truth label $\eta(i)$, we use exact methods to generate the global optimal scheduling results. RL pre-scheduling agent aims to imitate the behavior of the ground truth label. 

\noindent


\textbf{Training dataset - } Collecting sufficient data for Reinforcement Learning (RL) model training is one of the most significant challenges. The DNN computational graphs is limited, which restricts the scalability of RL agents. This challenge is exacerbated by the need to balance dataset coverage and graph size. To overcome this issue, we employ randomly generated synthetic datasets to train our RL model, which offers better data coverage over graph complexity and size while avoiding the need for extensive data collection. To ensure the generated DAGs align with real-world distributions, we gather statistical data from real-world DNN models and analyze the key information including graph size and complexities (maximum in-degree for nodes in the computational graph), etc. Building on this insight, we introduce a DAG generator tailored for RL training. This generator consistently outputs computational graphs with a size of $|V|=30$ and various graph complexities ranging from $deg(V) \in {2,3,4,5,6}$, mirroring the distributions found in real-world scenarios. Our DAG generator provides a total of \textbf{1 million} computational graphs, with 200,000 graphs for each complexity $deg(V)$ in ${2,3,4,5,6}$.

The synthetic dataset has superior generalizability with test datasets for two main reasons. Firstly, it mimics the graph structure and provides a similar degree of the test dataset, ensuring that the RL model is trained on synthetic data that closely resembles the test data. Secondly, the graph sampler generates a vast dataset of over 1 million synthetic graphs, providing extensive coverage of different types of test graphs. This large and diverse dataset improves the robustness and performance in real-world scenarios.



\subsection{Phase 2: Search Space Relaxation}

The main drawback of coarse-grained scheduling results from RL pre-scheduling is lacking determinism and optimality. In this section, we perform relaxation to broaden the search space and obtain the partial optimum. Space relaxation enables the application of exact methods to produce partial optimums without considering about scalability issues.

\begin{figure}[!htb]
     \centering
     \includegraphics[width=0.48\textwidth]{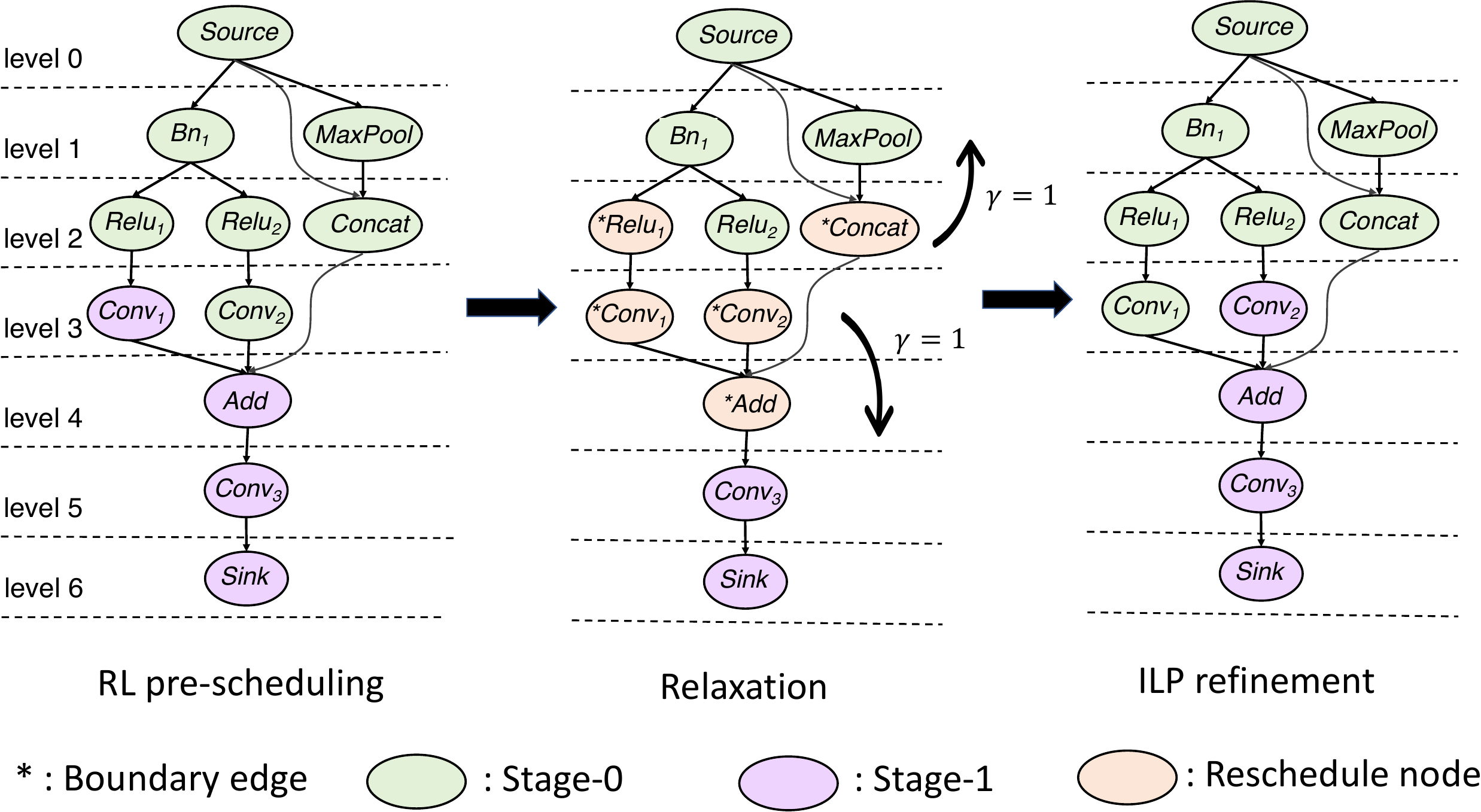}
     \caption{Search space relaxation based graph topological sorting.}
     \label{fig:relaxation}
 \end{figure}

Figure \ref{fig:relaxation} illustrates the search space relaxation of 2-stage pipelining with the relaxation level $\gamma = 1$. Firstly, RL pre-scheduling results pipeline the input model into 2 stages, where all boundary edges cut through stage-0 to stage-1. Second, we create an ASAP (As Soon As Possible) topological sort for the computation graph, positioning each node as near to the source node as feasible. The ASAP topological sorting can be seen on the left side of Figure \ref{fig:relaxation}. Thirdly, we find the earliest stage $s_i$ of the source node and the latest stage $s_j$ of the sink node in the boundary edges. Finally, we refine all nodes whose topological sorting order is located between $s_i$ to $s_j$. Using the computational graph in Figure \ref{fig:relaxation} as an example, the boundary edges include \{Relu\_1, Conv\_1\}, \{Conv\_2, Add\}, and \{Concat, Add\}. We use the star marker to identify the boundary edges. The earliest topological sorting level for source nodes is 2 (Relu\_1, Concat). Similarly, the latest sorting level for sink nodes is 4 (Add). Because we assume $\gamma=1$, all nodes whose sorting stage is located between 1 to 5 will be refined in the next step using exact methods. \textbf{While the exact methods based scheduling problem is an NP-hard problem, the relaxed problem on top of RL-based scheduling is significantly simplified due to drastic search space reduction.}


\subsection{Phase 3: Exact Solving via Multi-Obj ILP}\label{sec:ILP}


Phase 3 utilizes SDC+ILP-based scheduling \cite{zhang2013sdc} to execute the final refinement solving, aiming for optimal solutions. According to the systematic design of our multi-stage Edge TPU pipelining system, we adopt the following three optimization objectives when constructing ILP formulations: (1) peak memory usage for parameter caching; (2) total off-cache memory across all stages; (3) maximum communication cost in the multi-stage pipelining system. This section discusses the development of the ILP formulation. 

\noindent




\noindent
\textbf{ILP formulation generation}  --
In order to reschedule the refined search space, we construct an Integer Linear Programming (ILP) formulation comprised of various constraints. This formulation includes data dependence constraints, parameter caching constraints, and communication cost constraints, among others. In this section, we provide a concise overview of these three representative constraints.

\noindent
\textit{(1) Dependence constraint}  -- Dependence constraint formulates the dependence correctness of operators in the computation graph. Specifically, for edge $(N_i, N_j)$, the dependence constraint requires the $N_i$ to be executed before $N_j$. We use $s_i$ and $s_j$ to represent the stage of node $N_i, N_j$. Then the dependence constraint\cite{zhang2013sdc} can be formulated as:

\begin{center}
\footnotesize
\vspace{-3mm}
\begin{equation}
s_{\text{i}} - s_{\text{j}} \le 0
\label{eq:Dependence_constraints}
\end{equation}
\end{center}

\noindent
\textit{(2) Parameter caching constraint}  -- The most important optimization objective is the peak memory footprint $m_{peak}$. We use $m_k$ to denote the memory usage in stage k, which can be estimated by the sum of parameter sizes of operators that are scheduled to stage k. We use the following expression to formulate the parameter caching constraint:

\begin{center}
\footnotesize
\vspace{-3mm}
\begin{equation}
m_\text{k}=\sum_{\forall v \in V} p_v \cdot s_v^{k}
\label{eq:memory_cost}
\end{equation}
\end{center}

where variable $p_v$ is the estimation of memory consumption for operator $v$ based on parameter size, and $s_v^k$ is binary variable to denote if operator $v$ is scheduled to stage $k$.

$m_{peak}$ is the highest memory footprint among all stages:

\begin{equation}
\small
\centering
    \left\{
     \begin{array}{lr}
     \textbf{\text{min}} ~~m_{\text{peak}} \\
     \wedge ~~~~m_\text{k} \leq m_{\text{peak}} \\
     \wedge ~~~~\text{Equations (3)}
     \end{array}
         \right.\
         \label{eq:min_mlimit}
\end{equation}


\noindent
\textit{(3) Communication cost optimization} -- Communication cost constraints build ILP formulations to minimize the communication cost (i.e., tensor size) $com_{k,k+1}$ transferred between stages $k$ and stage $k+1$. For edge the $e$ from node $v_i$ to node $v_j$, we build the following ILP formulations:

\vspace{-2mm}

\begin{equation}
\footnotesize
     com_{k,k+1}=\sum_{e} t_\text{e} \cdot \alpha_\text{e}
\label{communication expression}
\end{equation}

where $t_\text{e}$ represents the tensor size between nodes $v_i$ and $v_j$, and $\alpha_\text{e}$ is a binary variable to denote if $(s_i==k) \wedge (s_i < s_j)$. 

In the previous formulation, we employ logical operators to represent the variable $\alpha_\text{e}$. Consequently, we require an additional set of constraints to represent these logical operators. For instance, consider the logical AND operation. Given two binary variables $x$ and $y$, the logical AND $\text{ILP}_{\wedge}(x,y)$ can be represented as follows:

\vspace{-4mm}

\begin{center}
\small
\begin{equation}
\left\{
     \begin{array}{lr}
     \text{ILP}_{\wedge}(x,y) \ge x+y-1\\
     \text{ILP}_{\wedge}(x,y) \le x\\
     \text{ILP}_{\wedge}(x,y) \le y\\
     \end{array}
\right.
\label{eq:and}
\end{equation}
\end{center}

\subsection{Phase 4: Deployment and system integration}


\vspace{-4mm}

 \begin{figure}[!htb]
  \centering
  \begin{subfigure}[b]{0.30\textwidth}
  \centering
  \includegraphics[width=1\linewidth]{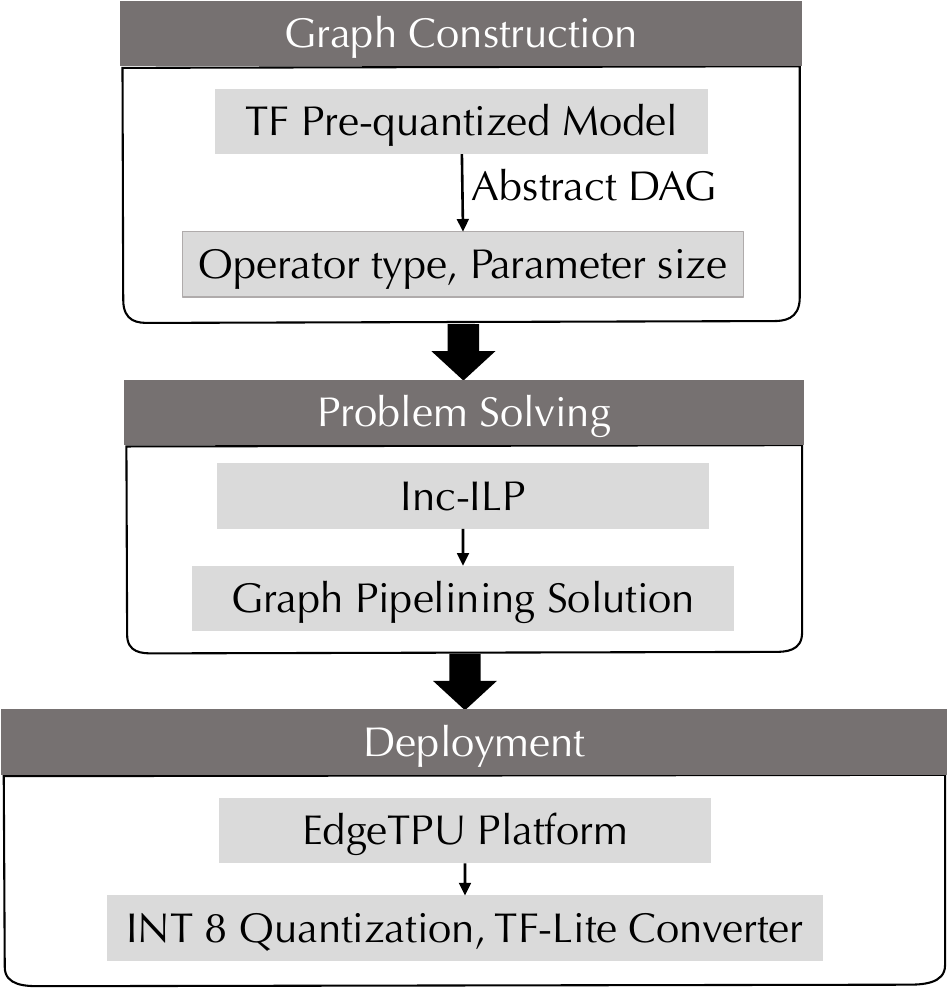}
  \caption{Inc-ILP overflow}
  \label{fig:deployment overflow}
  \end{subfigure}
  \begin{subfigure}[b]{0.30\textwidth}
  \centering
  \includegraphics[width=1\linewidth]{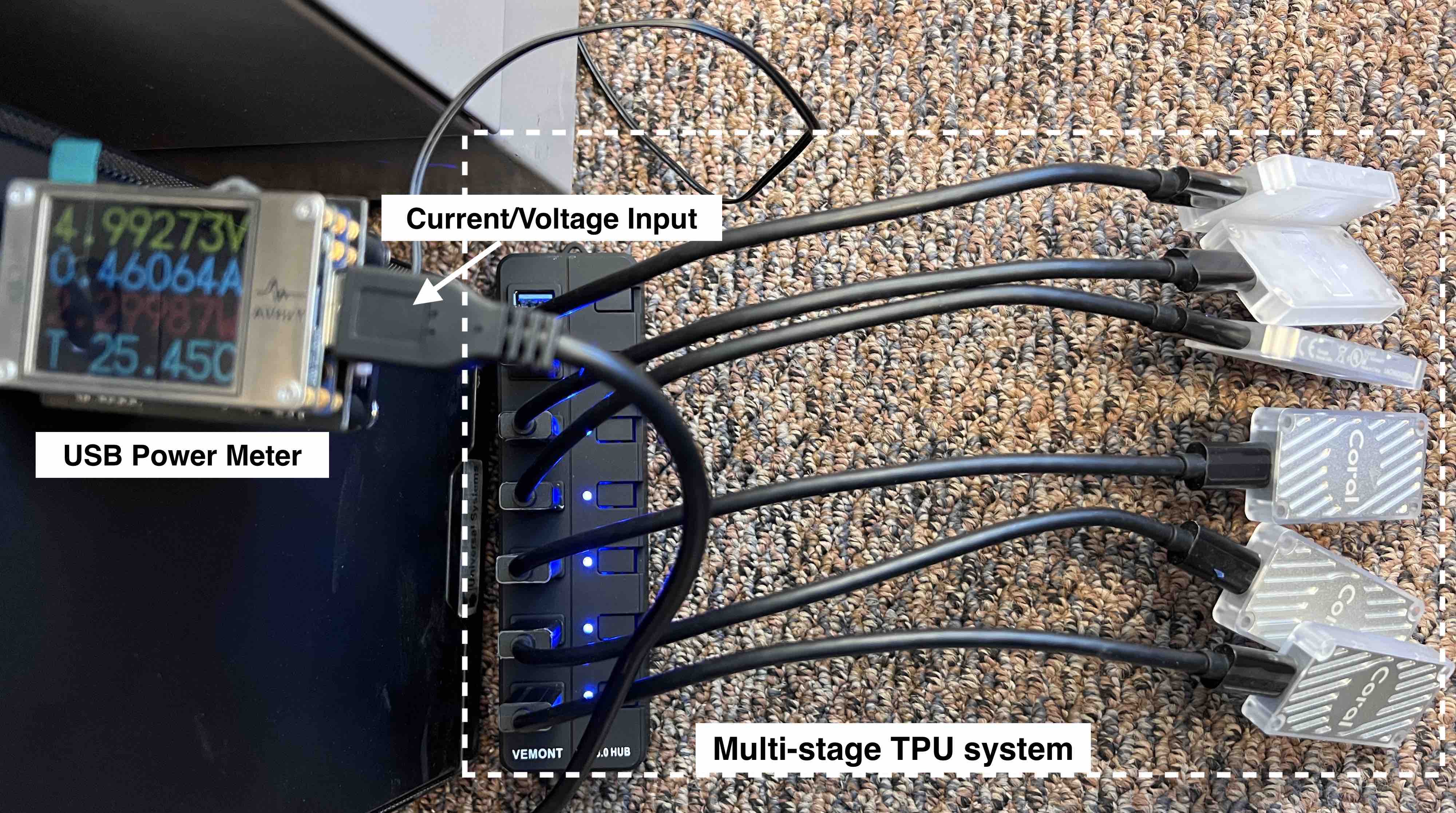}
  \caption{Multi-stage pipeline Edge TPUs system}
  \label{gap_5_stage}
  \label{fig:edgetpu_system}
  \end{subfigure}
  \caption{Inc-ILP end-to-end overflow and multi-stage pipeline Edge TPUs and energy measurement system.}
  \label{fig:implementation}
\end{figure}

\begin{figure*}[!htb]
  \centering
  \begin{subfigure}[t]{0.49\textwidth}
  \raisebox{-\height}{\includegraphics[width=\textwidth]{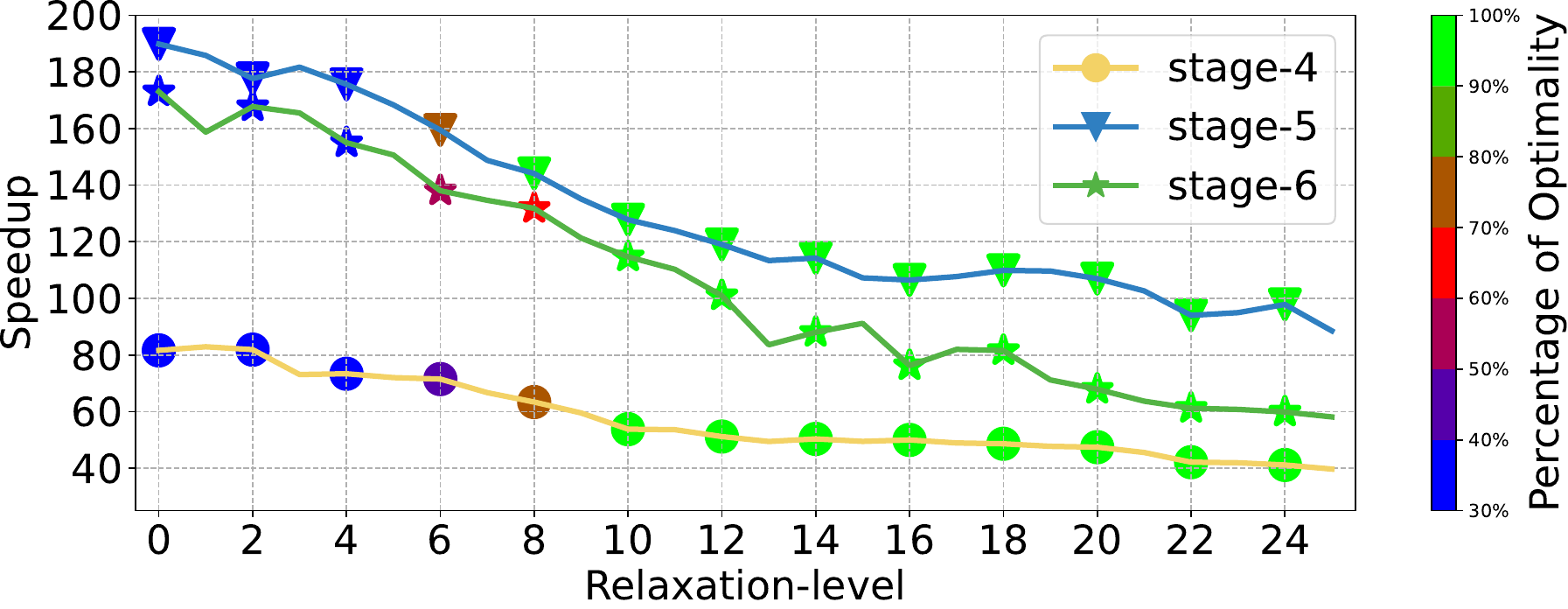}}
  \caption{Speedup over exact methods}
  \end{subfigure}
  \hfill
  \begin{subfigure}[t]{0.49\textwidth}
  \raisebox{-\height}{\includegraphics[width=\textwidth]{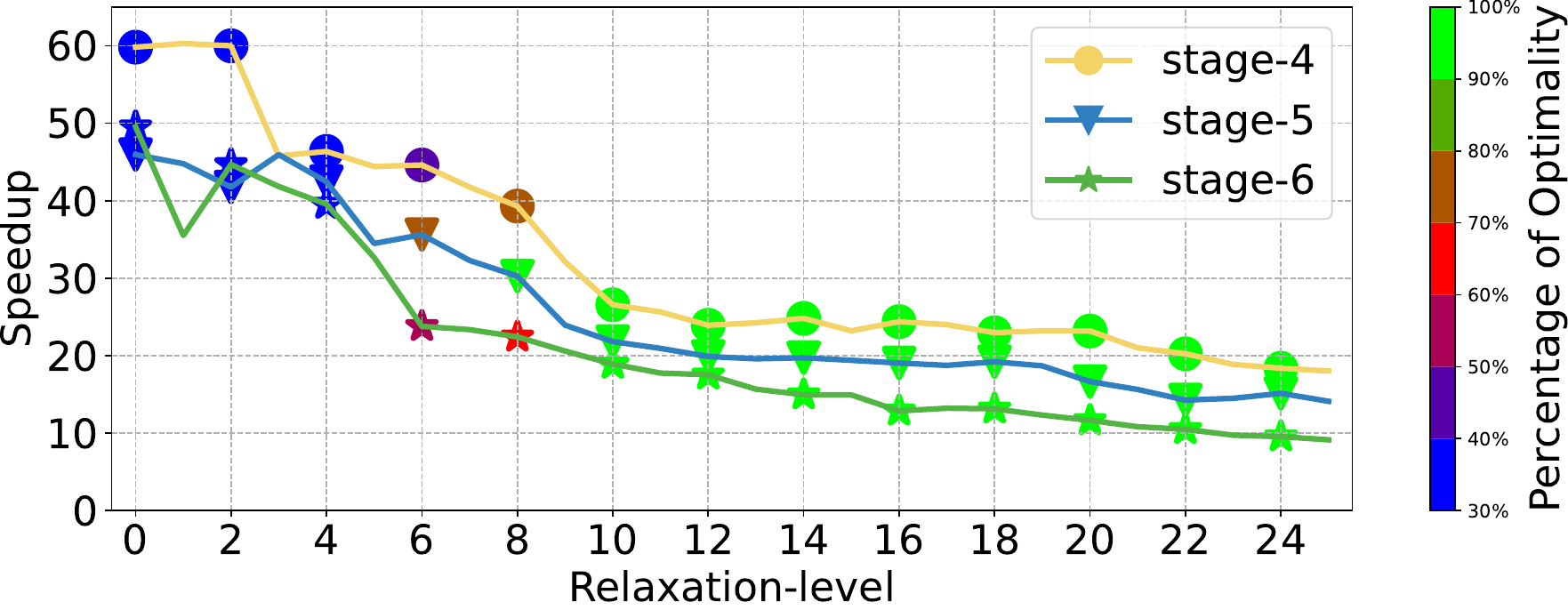}}
  \caption{Speedup over Edge TPU compiler}
  \end{subfigure}
  \caption{Inc-ILP solving runtime speedup over two baselines. According to the colorbar, the percentage of benchmarks reaching optimum is shown in marker color. Inc-ILP provides global optimal scheduling results with significant solving runtime speedup.}
  \label{fig:speedup}
\end{figure*}
 
We build an on-device framework that integrates the Inc-ILP with multi-stage pipelined Edge TPU systems and the illustration of the integration framework is shown in Figure \ref{fig:implementation}. The workflow includes the following several steps:

\begin{itemize}[leftmargin=*]
    \item \textbf{Graph construction - } 
    The Inc-ILP approach takes a graph, representing the dataflow of deep neural networks (DNNs), as its initial input. Our framework initially extracts the computation graph from TensorFlow Lite (TFLite), which includes details such as operator type, parameter dimensions, and output volume size, among others.
    
    \item \textbf{Scheduling via Inc-ILP - }
    With the computational graph, Inc-ILP solves the formulations generated in the prior step and extracts solutions corresponding to the original model(s) TensorFlow frozen graph. Subsequently, our framework employs the TFLite TOCO converter to generate model subgraphs, which will be assigned to particular Edge TPU devices based on the obtained solution.
    \item \textbf{Deployment - } 
    In the final step, our framework utilizes the Edge TPU Compiler to transform the subgraph operations into Edge TPU-specific operations and deploy the subgraphs onto Edge TPUs without any additional optimizations. As all Edge TPU models undergo INT8 quantization prior to deployment, our scheduling framework factors in this quantization when generating communication and memory caching constraints.
    
\end{itemize}

\vspace{-2mm}

\begin{table}[!h]
\scriptsize
    \captionof{table}{Statistics of DNN computational graph benchmarks}
    \label{tbl:dnn_graphs}
    \resizebox{0.45\textwidth}{!}{%
    \begin{tabular}{|c|c|c|c|c|}\hline
          & \begin{tabular}{@{}c@{}} Xception \\ \cite{chollet2017xception}\end{tabular} & \begin{tabular}{@{}c@{}} ResNet50 \\ \cite{he2016deep}\end{tabular} & \begin{tabular}{@{}c@{}} ResNet101 \\ \cite{he2016deep}\end{tabular} & \begin{tabular}{@{}c@{}} ResNet152 \\ \cite{he2016deep}\end{tabular} \\\hline
        $\mathbf{|V|}$ & 134 & 177 & 347 & 517\\\hline
        $\mathbf{deg(V)}$ & 2 & 2 & 2 & 2\\\hline
        \textbf{Depth} & 125 & 168 & 338 & 508\\\hline
         & \begin{tabular}{@{}c@{}} DenseNet121 \\ \cite{huang2017densely}\end{tabular} & \begin{tabular}{@{}c@{}} ResNet101v2 \\ \cite{he2016identity}\end{tabular} & \begin{tabular}{@{}c@{}} ResNet152v2 \\ \cite{he2016identity}\end{tabular} & \begin{tabular}{@{}c@{}} DenseNet169 \\ \cite{huang2017densely}\end{tabular} \\\hline
        $\mathbf{|V|}$ & 429 & 379 & 566 & 597\\\hline
        $\mathbf{deg(V)}$ & 2 & 2 & 2 & 2\\\hline
        \textbf{Depth} & 428 & 371 & 558 & 596 \\\hline
          & \begin{tabular}{@{}c@{}} DenseNet201 \\ \cite{huang2017densely}\end{tabular} & \begin{tabular}{@{}c@{}} InceptionResNetv2 \\ \cite{szegedy2017inception}\end{tabular} & & \\\hline
        $\mathbf{|V|}$ & 709 & 782 & & \\\hline
        $\mathbf{deg(V)}$ & 2 & 4 & & \\\hline
        \textbf{Depth} & 708 & 571 &  & \\\hline
    \end{tabular}}
    \label{DNNs}
\end{table}

\vspace{-3mm}

\section{Experiment}

In this section, we aim to experimentally demonstrate the results and effectiveness of Inc-ILP results. The comparison baselines include exact method (ILP) and the heuristic method (commercial EdgeTPU compiler). We demonstrate the performance of our work from three aspects: quality of results w.r.t parameter caching, solving runtime speedup, and on-device runtime performance.

\subsection{Experiment Setup}

We select Google Edge TPU as Inc-ILP backends and choose 10 ImageNet classification models as benchmarks to verify the effectiveness of Inc-ILP which is shown in Table \ref{tbl:dnn_graphs}. During the RL model training process, we use Nvidia 2080 Ti with Intel Xeon Gold 6230 x20 CPUs. For RL training setup, we select $10^{-4}$ as the learning rate with 300 epochs and batch size 128 using Adam optimizer. As we mentioned before, the RL encoder and decoder are built with 256 LSTM blocks. Besides, we adopt the IBM ILOG CPLEX \footnote{https://www.ibm.com/analytics/cplex-optimizer} as the incremental ILP solver. Finally, we select the Edge TPU compiler as the baseline which is a heuristic-based DNN scheduling commercial tool for Edge TPU devices. We build a multi-stage Edge TPU system for on-device evaluation which is shown in Figure \ref{fig:edgetpu_system}.

\vspace{-1mm}
\subsection{Quality of results and solving runtime analysis}


\label{sec:memory_gap}




The main goal of Inc-ILP is to achieve optimal solutions at significantly reduced runtime. Thus, we thoroughly examine its capacity to achieve optimal solutions and analyze the runtime expenses involved in the process.


\begin{figure*}[!htb]
    \centering
    \begin{subfigure}[t]{1\textwidth}
        \raisebox{-\height}{\includegraphics[width=\textwidth]{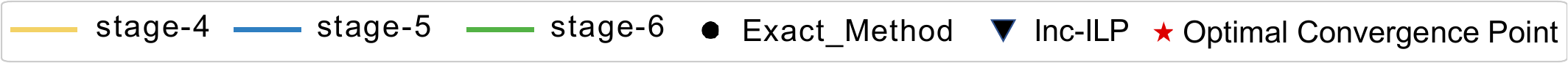}}
    \end{subfigure}
    
    \vspace{1mm}

     \centering
    \begin{subfigure}[t]{0.245\textwidth}
        \raisebox{-\height}{\includegraphics[width=\textwidth]{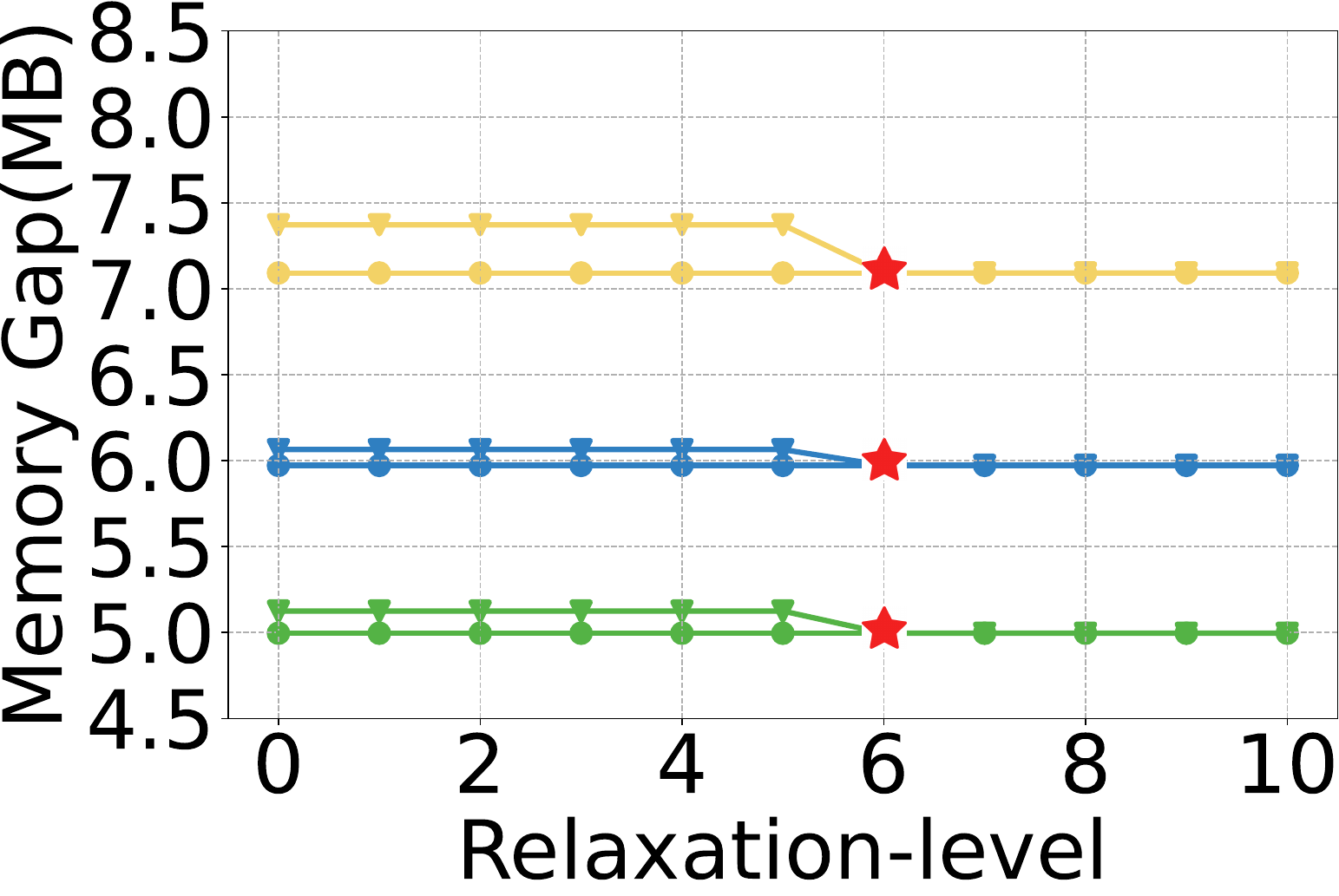}}
        \caption{DenseNet121}
    \end{subfigure}
    \hfill
    \begin{subfigure}[t]{0.245\textwidth}
        \raisebox{-\height}{\includegraphics[width=\textwidth]{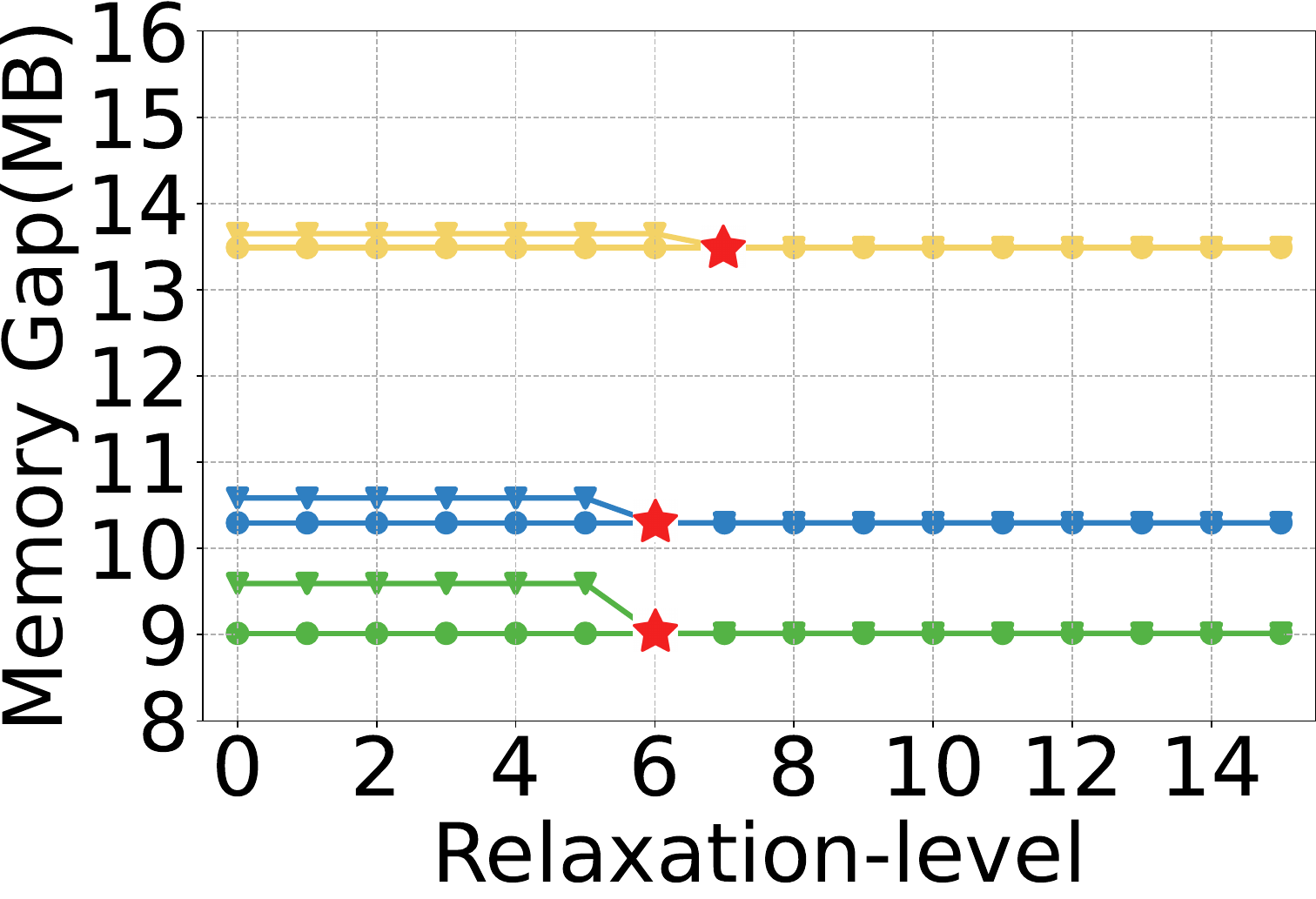}}
        \caption{DenseNet169}
    \end{subfigure}
    \hfill
    \begin{subfigure}[t]{0.245\textwidth}
        \raisebox{-\height}{\includegraphics[width=\textwidth]{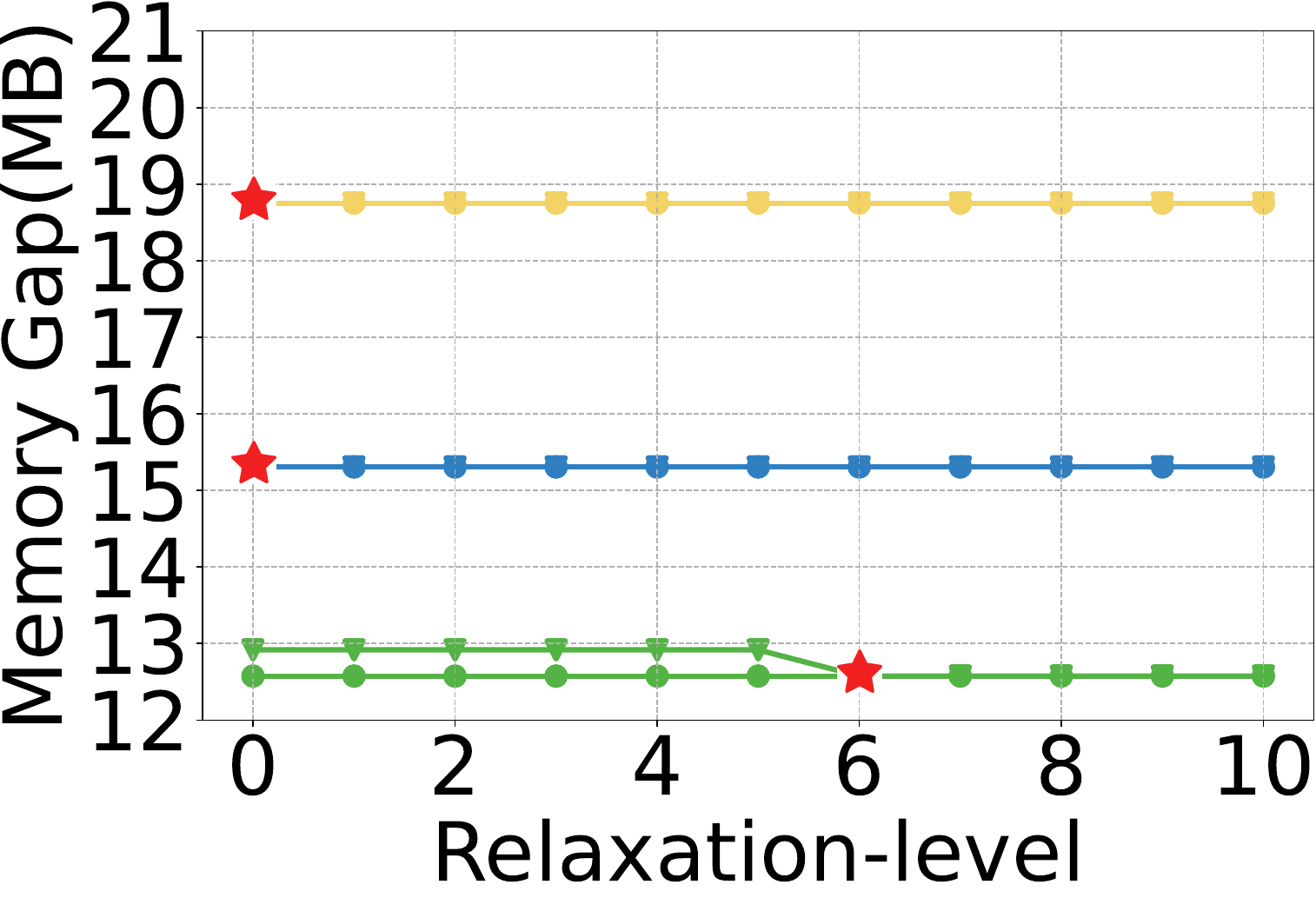}}
    \caption{DenseNet201} 
    \end{subfigure}
    \hfill
    \begin{subfigure}[t]{0.245\textwidth}
        \raisebox{-\height}{\includegraphics[width=\textwidth]{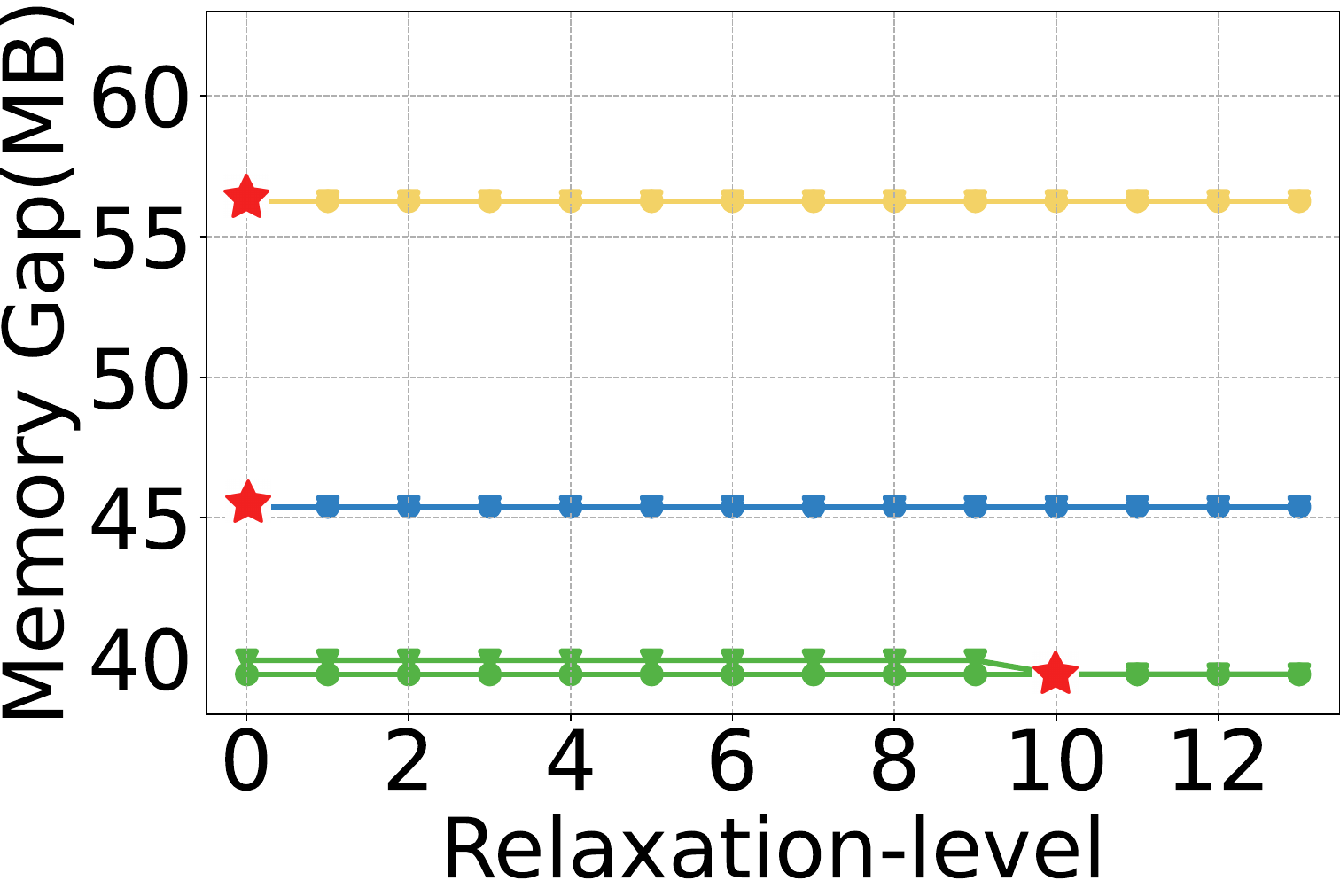}}
        \caption{Inception\_ResNet\_v2}
    \end{subfigure}

    \begin{subfigure}[t]{0.245\textwidth}
        \raisebox{-\height}{\includegraphics[width=\textwidth]{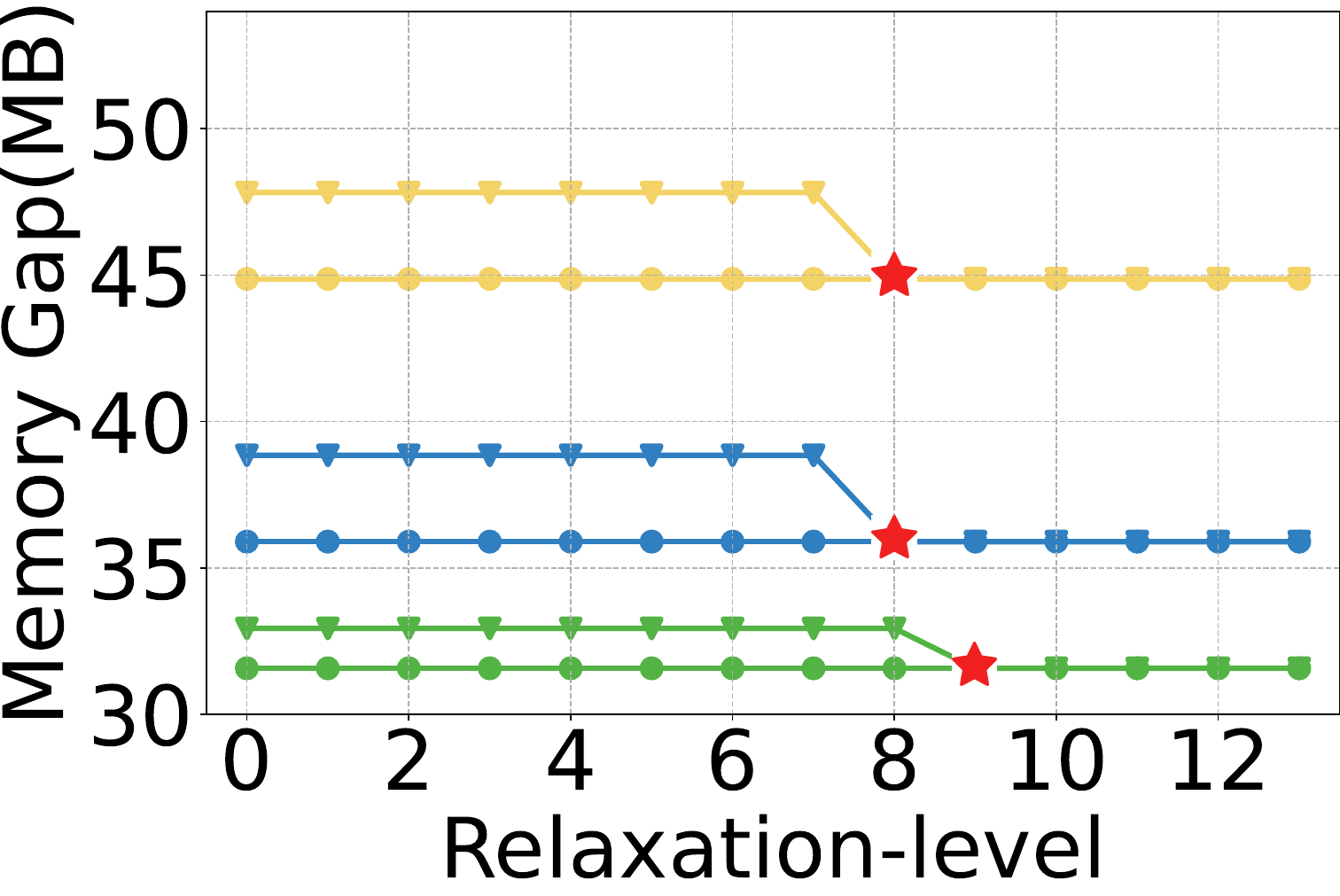}}
    \caption{ResNet101} 
    \end{subfigure}
    \hfill
    \begin{subfigure}[t]{0.245\textwidth}
        \raisebox{-\height}{\includegraphics[width=\textwidth]{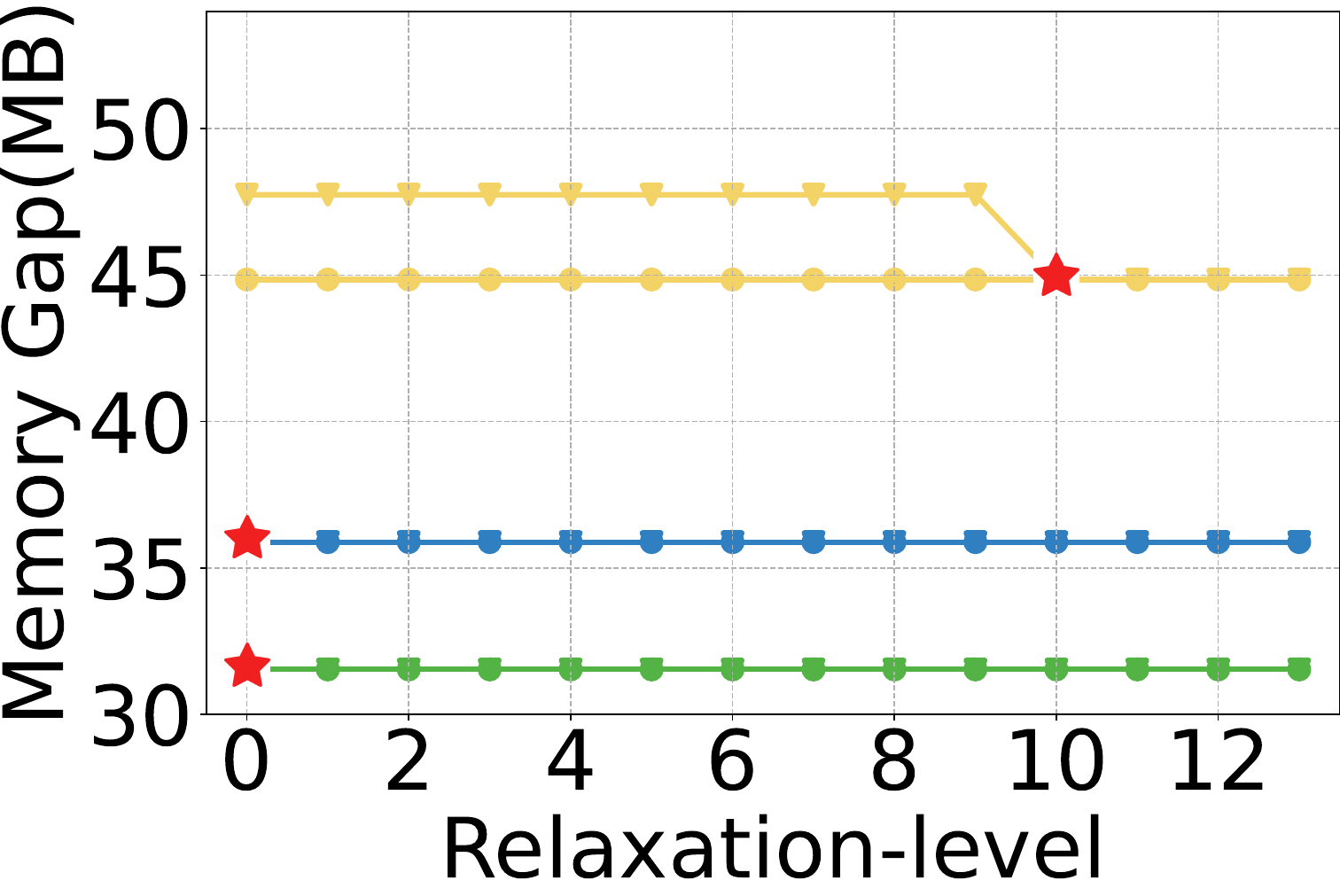}}
        \caption{ResNet101v2}
    \end{subfigure}
    \hfill
    \begin{subfigure}[t]{0.245\textwidth}
        \raisebox{-\height}{\includegraphics[width=\textwidth]{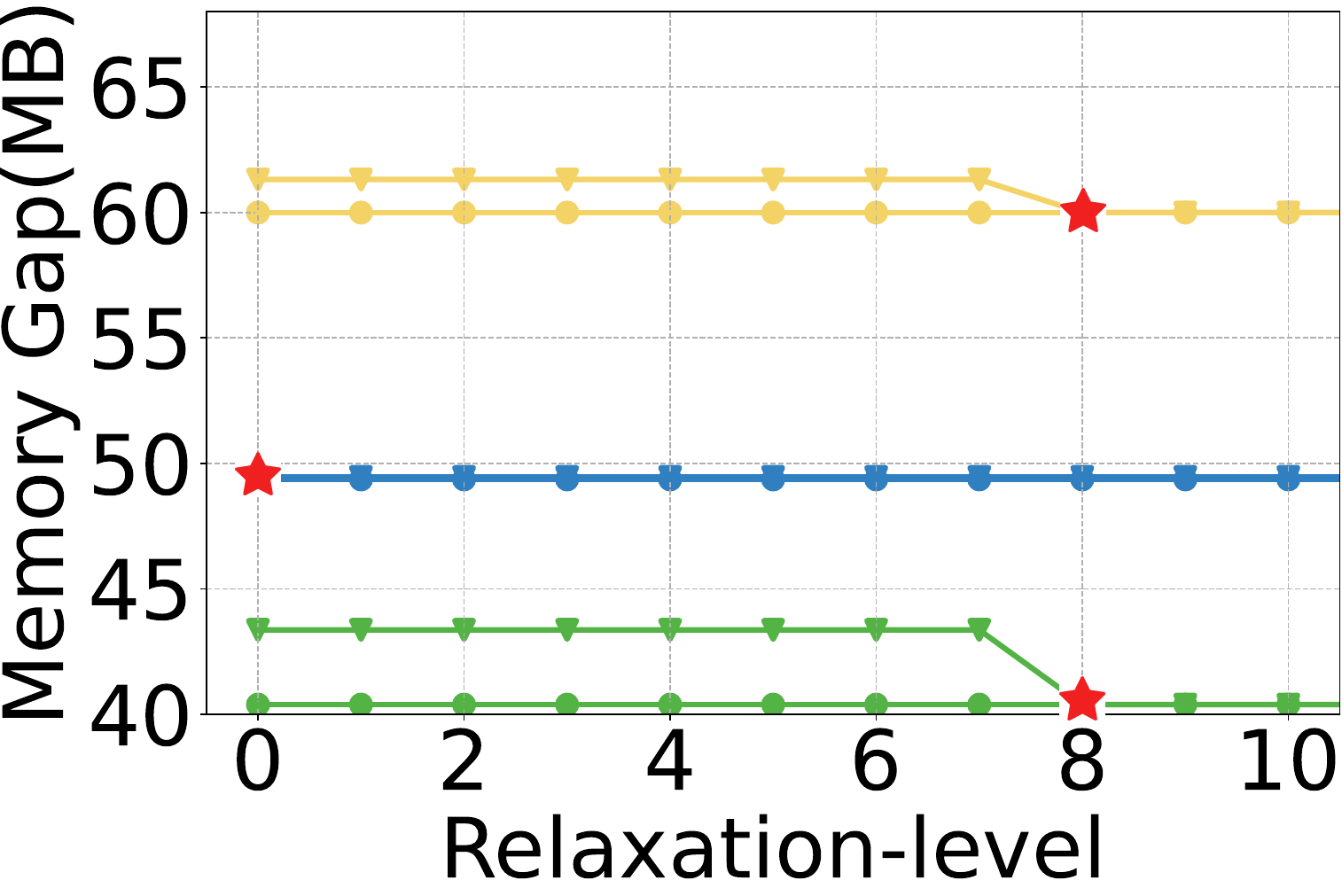}}
    \caption{ResNet152} 
    \end{subfigure}

    \begin{subfigure}[t]{0.245\textwidth}
        \raisebox{-\height}{\includegraphics[width=\textwidth]{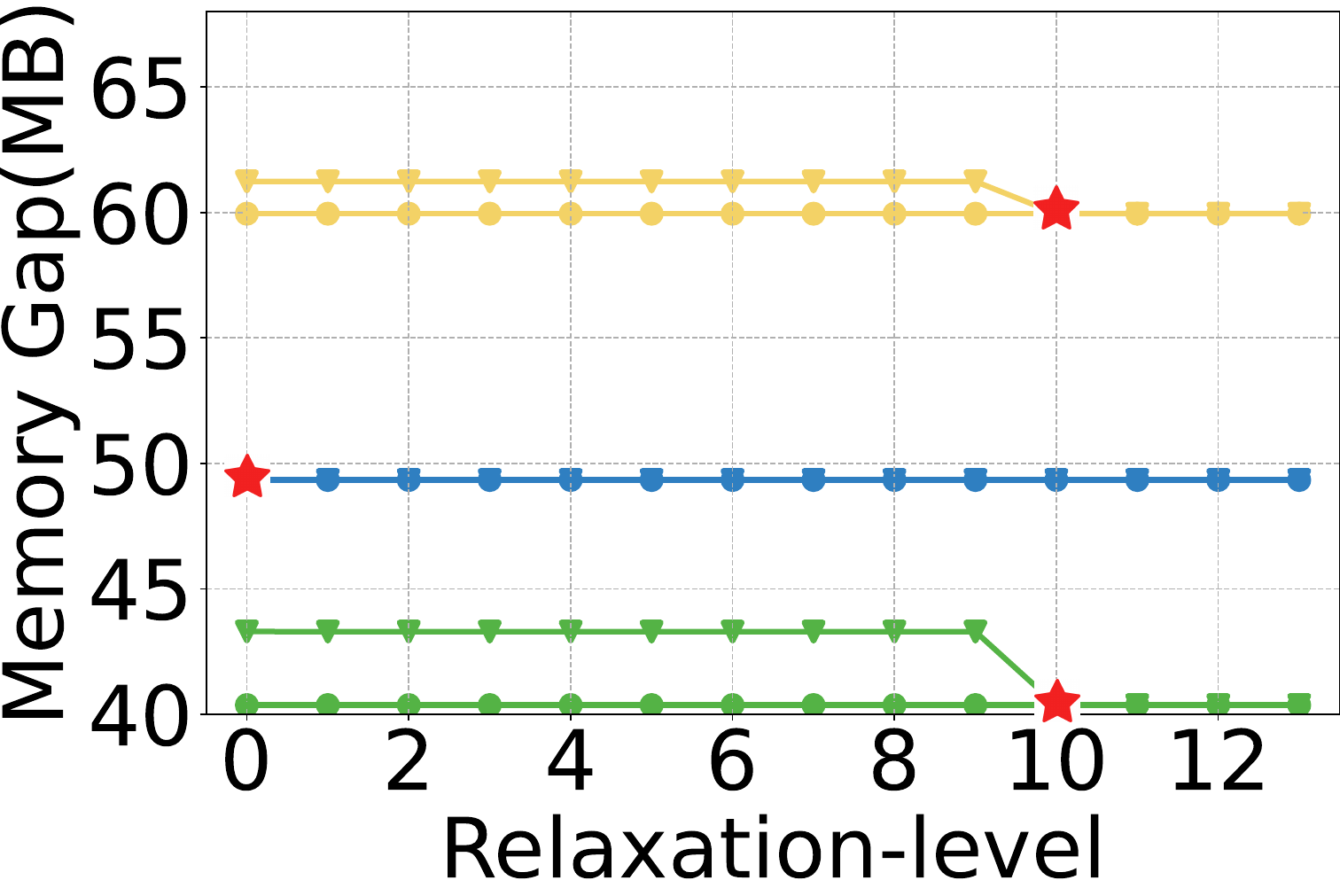}}
        \caption{ResNet152v2}
    \end{subfigure}
    \hfill
    \begin{subfigure}[t]{0.245\textwidth}
        \raisebox{-\height}{\includegraphics[width=\textwidth]{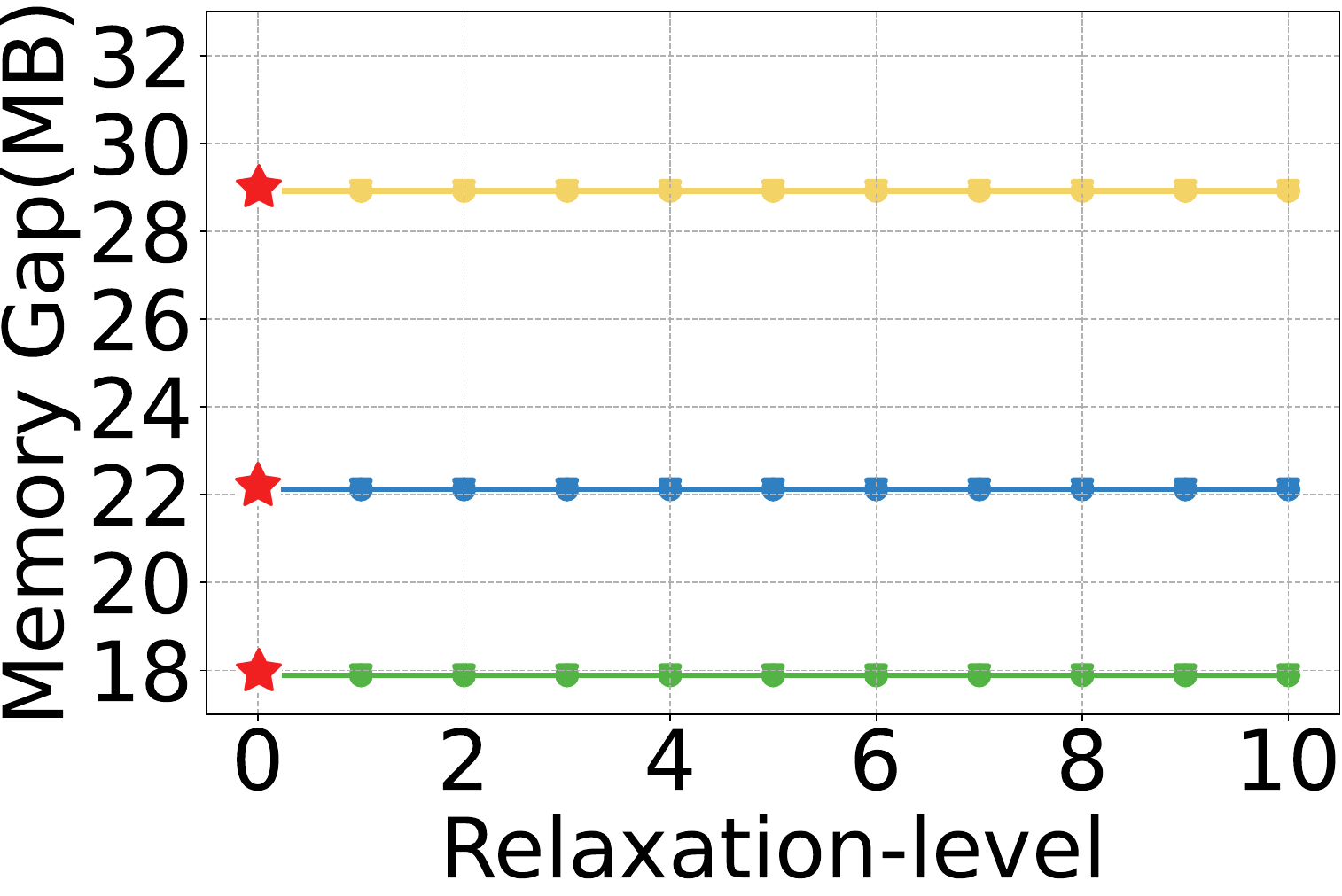}}
    \caption{ResNet50} 
    \end{subfigure}
    \hfill
    \begin{subfigure}[t]{0.245\textwidth}
        \raisebox{-\height}{\includegraphics[width=\textwidth]{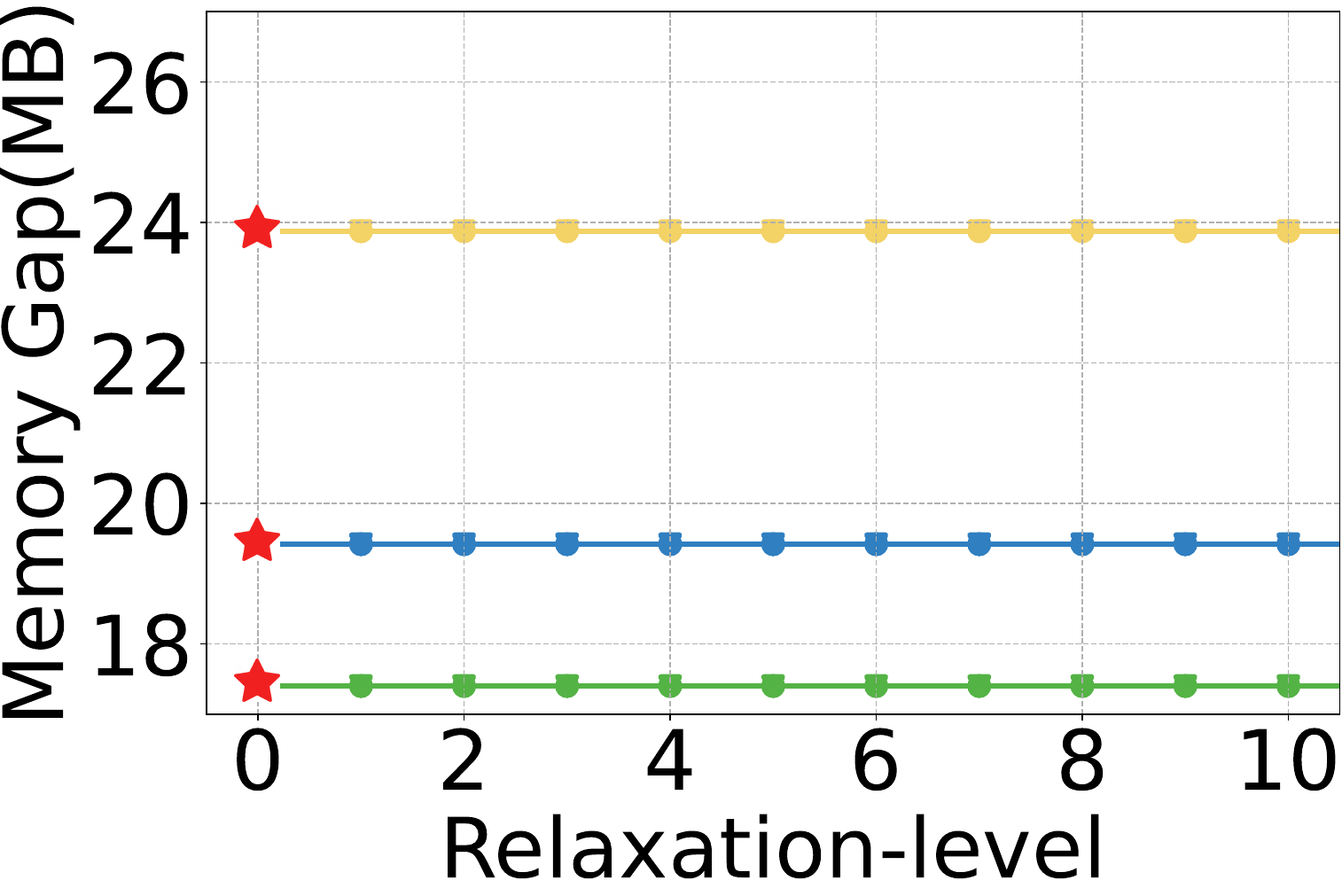}}
        \caption{Xception}
    \end{subfigure}

    \caption{Convergence analysis w.r.t relaxation depth using parameter caching metric as example. All three objectives reach optimal after the convergence point.}
    \label{fig:Mem_Gap}
\end{figure*}



First, we compare the solving runtime speedups of Inc-ILP with two baselines: ILP-based exact method and commercial Edge TPU compiler, while the quality-of-result metrics if the percentage of DNNs graphs are optimally scheduled (Figure \ref{fig:speedup}). The vertical axis is the Inc-ILP average solving runtime speedup over all ten benchmarks in Table \ref{tbl:dnn_graphs}, where the percentage of benchmarks that achieve global optimum are presented w.r.t the color-bar. 
As shown in Figure \ref{fig:speedup}, Inc-ILP offers significant solving runtime speedup over exact method with identical optimal solutions, where full optimality (100\%) is reached at relaxation level $\gamma=10$. Specifically, Inc-ILP provides $54 \times$, $128 \times$, and $115 \times$ speedups for 4-stage, 5-stage, and 6-stage scheduling, respectively ($\gamma=10$), with full optimality achieved. This confirms that RL-based initialization successfully narrows the search space, which alleviates the scalability issue of exact methods and makes finding the optimal solution within short solving runtime possible. Secondly, Inc-ILP offers $27\times$, $22\times$, and $19\times$ speedups ($\gamma=10$) over EdgeTPU compiler (heuristic), while offering guarantees in optimality. In conclusion, Inc-ILP not only provides the solving runtime improvement but also generates the optimal scheduling solution. We can also see that relaxation level $\gamma$ is an important factor that drives the trade-off between optimality and runtime. 

To understand the impacts of the relaxation level $\gamma$, we perform a case-by-case analysis for each benchmarks, shown in Figure \ref{fig:Mem_Gap}. While we use the peak parameter caching as the objective metric for this evaluation, the three metrics all reach optimal points after careful search space relaxation. In this figure, we compare the size of peak memory usage between Inc-ILP and exact methods where the $x$-axis and $y$-axis represent the relaxation level $\gamma$ and the parameter caching results of Inc-ILP and the optimum. 
Three important conclusions can be derived from Figure \ref{fig:Mem_Gap}. First, RL-based pre-scheduling provides a great initialization result for later exact ILP solving. For example, three benchmarks achieve the global optimum without ILP refinement using RL-based scheduling ($\gamma=0$). As for all benchmarks, the scheduling results from the RL pre-scheduling agent have peak parameter caching gaps of {1.88\%, 1.12\%, and 4.50\%} to optimum, in 4-stage, 5-stage, and 6-stage pipelining, respectively.   
The quality of the initialization confirms the great confidence in including the global optimal solution in the relaxed search space. Second, most benchmarks achieve optimal scheduling results with small relaxation levels. For example, ResNet152 depth is 508, while the optimal convergence point is reached by Inc-ILP with only $\gamma=8$. {We observe that for all graphs tested, the 10-level relaxation is sufficient to generate global optimal results for all cases. Increasing the relaxation level beyond 10 brings marginal benefits but could reduce the advantages of Inc-ILP in solving runtime speedups.} Finally, although the graph size of DNN computation graph benchmarks is vary from 134 to 782, the relaxation level to achieve optimal peak-memory results remains stable for different benchmarks, which proves the scalability and generalizability of Inc-ILP. For example, both DenseNet121 and DenseNet201 achieve optimal results when $\gamma=6$, but they differ in graph size ($V$=429 for DenseNet121 and $V$=709 for DenseNet201).


\vspace{-1mm}

\subsection{On-device inference runtime analysis}

\begin{figure}[!htb]
  \centering
  \begin{minipage}{0.5\textwidth}
    \centering
    \subcaptionbox{4-stage}
      {\includegraphics[width=1\linewidth]{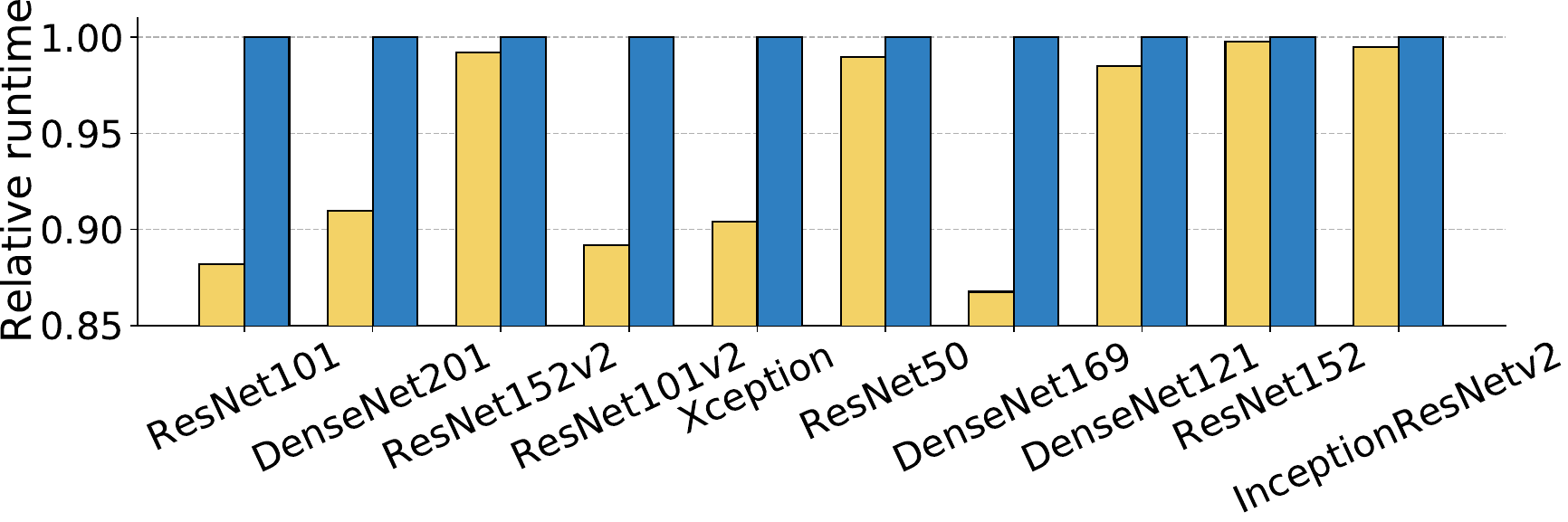}}
    \subcaptionbox{5-stage}
      {\includegraphics[width=1\linewidth]{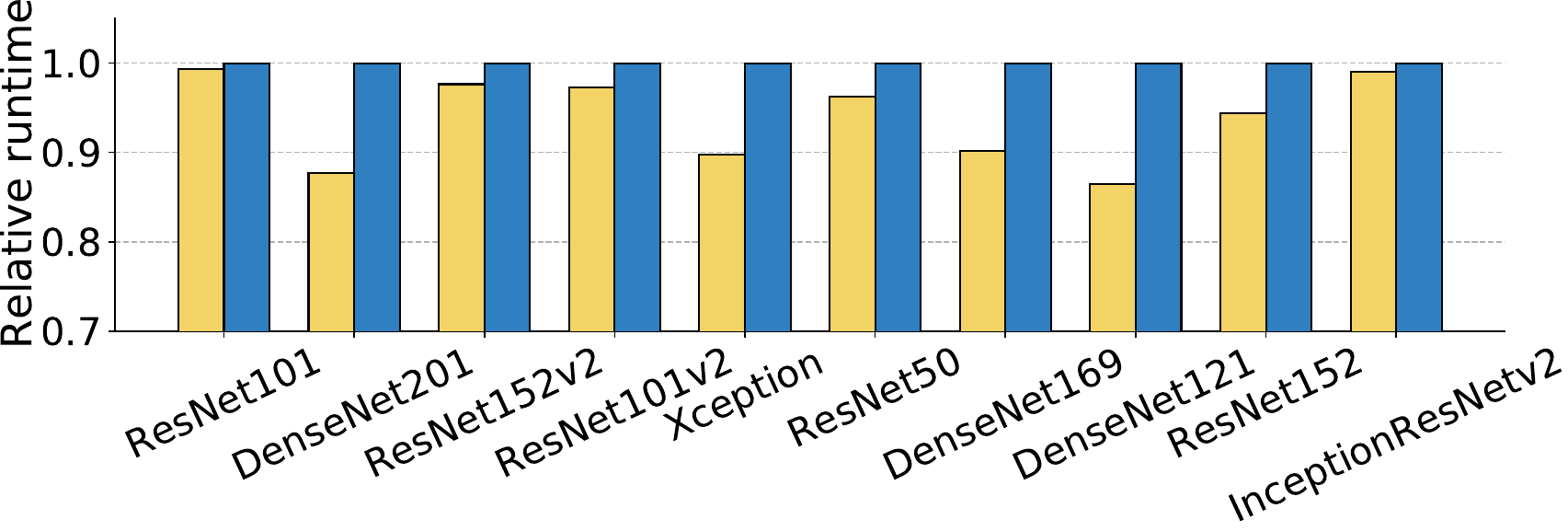}}
    \subcaptionbox{6-stage}
      {\includegraphics[width=1\linewidth]{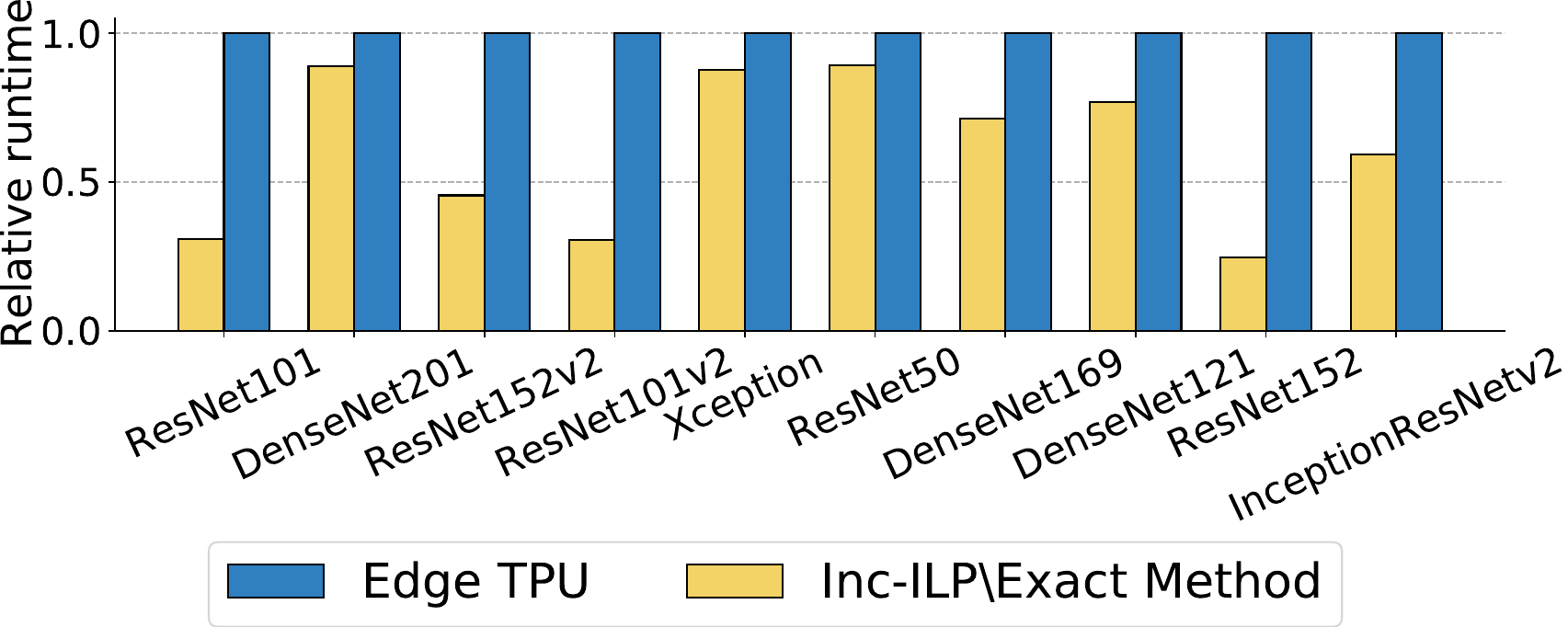}}
  \end{minipage}
  \caption{On-device runtime comparison (Inc-ILP meet same performance of exact method). Inc-ILP performs better than Edge TPU compiler for all benchmarks.}
  \label{fig:edgetpu_runtime}
\end{figure}

Finally, we compare the runtime performance between Inc-ILP, exact methods, and commercial Edge TPU compiler. The test system is shown in Figure \ref{fig:implementation} and the pipelined Edge TPU devices are connected to stable 5V supplying voltage for all tests. The on-chip inference runtime results are shown in Figure \ref{fig:edgetpu_runtime}. The vertical axis represents the relative runtime normalized to the Edge TPU compiler. Note that as we mentioned before, Inc-ILP has identical scheduling results with exact methods. Thus we merge the scheduling results of Inc-ILP and exact methods into one column.

As shown in Figure \ref{fig:edgetpu_runtime}, Inc-ILP has significant runtime performance improvement over the Edge TPU compiler. Inc-ILP benefits from the ILP refinement and guarantees partial optimality. Overall, Inc-ILP provides $1.065 \times$, $1.068 \times$, and $2.056 \times$ execution runtime performance speedup in 4-stage, 5-stage, and 6-stage pipelining. For example, Inc-ILP provides $4.07 \times$ runtime speedup for ResNet152 in 6-stage pipelining. Besides, we observe more substantial performance improvement in 6-stage pipelining. The main reason is the scheduling results quality degradation of heuristic methods when scheduling complexity increases for large number stage pipelining.

\section{Conclusion}

This work proposes a novel approach in enhancing ML-based optimization approaches in determinism and optimality. The key concept leverages ML-based solving at earlier stage that is coarse but low runtime cost, and deploys exact solving to refine the coarse solution to reach optimality from significantly reduced search space. Specifically, we propose a novel framework for scheduling, namely Inc-ILP, a DNN model scheduling framework that combines both exact methods and RL-based scheduling approaches. RL pre-scheduling agent narrows the searching space and Inc-ILP strategically refines RL initialization results with the exact method. Inc-ILP benefits from the scalability of the RL-based scheduling approach as well as the scheduling optimality of exact methods. The experimental results confirm the proposed concept, where Inc-ILP obtains the exact optimal scheduling with significant speedups over exact methods, and heuristic-based commercial EdgeTPU compiler (non-optimal scheduling).




\begin{thebibliography}{10}
\providecommand{\url}[1]{#1}
\csname url@samestyle\endcsname
\providecommand{\newblock}{\relax}
\providecommand{\bibinfo}[2]{#2}
\providecommand{\BIBentrySTDinterwordspacing}{\spaceskip=0pt\relax}
\providecommand{\BIBentryALTinterwordstretchfactor}{4}
\providecommand{\BIBentryALTinterwordspacing}{\spaceskip=\fontdimen2\font plus
\BIBentryALTinterwordstretchfactor\fontdimen3\font minus
  \fontdimen4\font\relax}
\providecommand{\BIBforeignlanguage}[2]{{%
\expandafter\ifx\csname l@#1\endcsname\relax
\typeout{** WARNING: IEEEtran.bst: No hyphenation pattern has been}%
\typeout{** loaded for the language `#1'. Using the pattern for}%
\typeout{** the default language instead.}%
\else
\language=\csname l@#1\endcsname
\fi
#2}}
\providecommand{\BIBdecl}{\relax}
\BIBdecl

\bibitem{jouppi2017datacenter}
N.~P. Jouppi, C.~Young, N.~Patil, D.~Patterson, G.~Agrawal, R.~Bajwa, S.~Bates,
  S.~Bhatia, N.~Boden, A.~Borchers \emph{et~al.}, ``In-datacenter performance
  analysis of a tensor processing unit,'' in \emph{ISCA}, 2017, pp. 1--12.

\bibitem{rittinghouse2017cloud}
J.~W. Rittinghouse and J.~F. Ransome, \emph{Cloud computing: implementation,
  management, and security}.\hskip 1em plus 0.5em minus 0.4em\relax CRC press,
  2017.

\bibitem{leiserson1991retiming}
C.~E. Leiserson and J.~B. Saxe, ``Retiming synchronous circuitry,''
  \emph{Algorithmica}, vol.~6, no.~1, pp. 5--35, 1991.

\bibitem{zhang2013sdc}
Z.~Zhang and B.~Liu, ``{SDC-Based Modulo Scheduling for Pipeline Synthesis},''
  \emph{Int'l Conf. on Computer-Aided Design (ICCAD)}, 2013.

\bibitem{he2016deep}
K.~He, X.~Zhang, S.~Ren, and J.~Sun, ``Deep residual learning for image
  recognition,'' in \emph{Proceedings of the IEEE conference on computer vision
  and pattern recognition}, 2016, pp. 770--778.

\bibitem{yin2022exact}
J.~Yin, Z.~Zhang, and C.~Yu, ``Exact memory-and communication-aware scheduling
  of dnns on pipelined edge tpus,'' in \emph{2022 IEEE/ACM 7th Symposium on
  Edge Computing (SEC)}.\hskip 1em plus 0.5em minus 0.4em\relax IEEE, 2022, pp.
  203--215.

\bibitem{yang1993list}
T.~Yang and A.~Gerasoulis, ``List scheduling with and without communication
  delays,'' \emph{Parallel Computing}, 1993.

\bibitem{ahn2020ordering}
B.~H. Ahn, J.~Lee, J.~M. Lin, H.-P. Cheng, J.~Hou, and H.~Esmaeilzadeh,
  ``Ordering chaos: Memory-aware scheduling of irregularly wired neural
  networks for edge devices,'' \emph{arXiv preprint arXiv:2003.02369}, 2020.

\bibitem{chen2018tvm}
T.~Chen \emph{et~al.}, ``{{TVM}: An automated end-to-end optimizing compiler
  for deep learning},'' in \emph{OSDI}, 2018, pp. 578--594.

\bibitem{abadi2016tensorflow}
M.~Abadi, P.~Barham, J.~Chen, Z.~Chen, A.~Davis, J.~Dean, M.~Devin,
  S.~Ghemawat, G.~Irving, M.~Isard \emph{et~al.}, ``Tensorflow: A system for
  large-scale machine learning,'' in \emph{12th $\{$USENIX$\}$ symposium on
  operating systems design and implementation ($\{$OSDI$\}$ 16)}, 2016, pp.
  265--283.

\bibitem{sanders2010cuda}
J.~Sanders and E.~Kandrot, \emph{CUDA by example: an introduction to
  general-purpose GPU programming}.\hskip 1em plus 0.5em minus 0.4em\relax
  Addison-Wesley Professional, 2010.

\bibitem{ren2023machine}
H.~Ren and J.~Hu, \emph{Machine Learning Applications in Electronic Design
  Automation}.\hskip 1em plus 0.5em minus 0.4em\relax Springer Nature, 2023.

\bibitem{huang2021machine}
G.~Huang, J.~Hu, Y.~He, J.~Liu, M.~Ma, Z.~Shen, J.~Wu, Y.~Xu, H.~Zhang,
  K.~Zhong \emph{et~al.}, ``Machine learning for electronic design automation:
  A survey,'' \emph{ACM Transactions on Design Automation of Electronic Systems
  (TODAES)}, vol.~26, no.~5, pp. 1--46, 2021.

\bibitem{yu2019painting}
C.~Yu and Z.~Zhang, ``Painting on placement: Forecasting routing congestion
  using conditional generative adversarial nets,'' in \emph{Proceedings of the
  56th Annual Design Automation Conference 2019}, 2019, pp. 1--6.

\bibitem{yu2018developing}
C.~Yu, H.~Xiao, and G.~De~Micheli, ``Developing synthesis flows without human
  knowledge,'' in \emph{Proceedings of the 55th Annual Design Automation
  Conference}, 2018, pp. 1--6.

\bibitem{wu2023gamora}
N.~Wu, Y.~Li, C.~Hao, S.~Dai, C.~Yu, and Y.~Xie, ``Gamora: Graph learning based
  symbolic reasoning for large-scale boolean networks,'' \emph{Design
  Automation Conference (DAC'23)}, 2023.

\bibitem{yu2020flowtune}
C.~Yu, ``Flowtune: Practical multi-armed bandits in boolean optimization,'' in
  \emph{Proceedings of the 39th International Conference on Computer-Aided
  Design}, 2020, pp. 1--9.

\bibitem{mao2019learning}
H.~Mao, M.~Schwarzkopf, S.~B. Venkatakrishnan, Z.~Meng, and M.~Alizadeh,
  ``Learning scheduling algorithms for data processing clusters,'' in
  \emph{Proceedings of the ACM Special Interest Group on Data Communication},
  2019, pp. 270--288.

\bibitem{chen2019deep}
H.~Chen and M.~Shen, ``A deep-reinforcement-learning-based scheduler for fpga
  hls,'' in \emph{2019 IEEE/ACM International Conference on Computer-Aided
  Design (ICCAD)}.\hskip 1em plus 0.5em minus 0.4em\relax IEEE, 2019, pp. 1--8.

\bibitem{sheng2021deep}
S.~Sheng, P.~Chen, Z.~Chen, L.~Wu, and Y.~Yao, ``Deep reinforcement
  learning-based task scheduling in iot edge computing,'' \emph{Sensors},
  vol.~21, no.~5, p. 1666, 2021.

\bibitem{yin2023respect}
J.~Yin, Y.~Li, D.~Robinson, and C.~Yu, ``Respect: Reinforcement learning based
  edge scheduling on pipelined coral edge tpus,'' \emph{arXiv preprint
  arXiv:2304.04716}, 2023.

\bibitem{boroumand2021mitigating}
A.~Boroumand \emph{et~al.}, ``Mitigating edge machine learning inference
  bottlenecks: An empirical study on accelerating google edge models,''
  \emph{arXiv preprint arXiv:2103.00768}, 2021.

\bibitem{williams1992simple}
R.~J. Williams, ``Simple statistical gradient-following algorithms for
  connectionist reinforcement learning,'' \emph{Machine learning}, vol.~8,
  no.~3, pp. 229--256, 1992.

\bibitem{bello2016neural}
I.~Bello, H.~Pham, Q.~V. Le, M.~Norouzi, and S.~Bengio, ``Neural combinatorial
  optimization with reinforcement learning,'' \emph{arXiv preprint
  arXiv:1611.09940}, 2016.

\bibitem{yu2020decision}
C.~Yu and W.~Zhou, ``Decision making in synthesis cross technologies using
  lstms and transfer learning,'' in \emph{Proceedings of the 2020 ACM/IEEE
  Workshop on Machine Learning for CAD}, 2020, pp. 55--60.

\bibitem{selsam2018learning}
D.~Selsam, M.~Lamm, B.~B{\"u}nz, P.~Liang, L.~de~Moura, and D.~L. Dill,
  ``Learning a sat solver from single-bit supervision,'' \emph{arXiv preprint
  arXiv:1802.03685}, 2018.

\bibitem{micheli1994synthesis}
G.~D. Micheli, \emph{{Synthesis and Optimization of Digital Circuits}}.\hskip
  1em plus 0.5em minus 0.4em\relax McGraw-Hill Higher Education, 1994.

\bibitem{fan2005cost}
K.~Fan, M.~Kudlur, H.~Park, and S.~Mahlke, ``Cost sensitive modulo scheduling
  in a loop accelerator synthesis system,'' in \emph{38th Annual IEEE/ACM
  International Symposium on Microarchitecture (MICRO'05)}.\hskip 1em plus
  0.5em minus 0.4em\relax IEEE, 2005, pp. 12--pp.

\bibitem{ramalingam1999solving}
G.~Ramalingam, J.~Song, L.~Joskowicz, and R.~E. Miller, ``{Solving Systems of
  Difference Constraints Incrementally},'' \emph{Algorithmica}, 1999.

\bibitem{dai2018scalable}
S.~Dai, G.~Liu, and Z.~Zhang, ``A scalable approach to exact
  resource-constrained scheduling based on a joint sdc and sat formulation,''
  in \emph{Proceedings of the 2018 ACM/SIGDA International Symposium on
  Field-Programmable Gate Arrays}, 2018, pp. 137--146.

\bibitem{hochreiter1997long}
S.~Hochreiter and J.~Schmidhuber, ``Long short-term memory,'' \emph{Neural
  computation}, vol.~9, no.~8, pp. 1735--1780, 1997.

\bibitem{vaswani2017attention}
A.~Vaswani, N.~Shazeer, N.~Parmar, J.~Uszkoreit, L.~Jones, A.~N. Gomez,
  {\L}.~Kaiser, and I.~Polosukhin, ``Attention is all you need,'' in
  \emph{Advances in neural information processing systems}, 2017, pp.
  5998--6008.

\bibitem{chollet2017xception}
F.~Chollet, ``Xception: Deep learning with depthwise separable convolutions,''
  in \emph{Proceedings of the IEEE conference on computer vision and pattern
  recognition}, 2017, pp. 1251--1258.

\bibitem{huang2017densely}
G.~Huang, Z.~Liu, L.~Van Der~Maaten, and K.~Q. Weinberger, ``Densely connected
  convolutional networks,'' in \emph{Proceedings of the IEEE conference on
  computer vision and pattern recognition}, 2017, pp. 4700--4708.

\bibitem{he2016identity}
K.~He, X.~Zhang, S.~Ren, and J.~Sun, ``Identity mappings in deep residual
  networks,'' in \emph{European conference on computer vision}.\hskip 1em plus
  0.5em minus 0.4em\relax Springer, 2016, pp. 630--645.

\bibitem{szegedy2017inception}
C.~Szegedy, S.~Ioffe, V.~Vanhoucke, and A.~A. Alemi, ``Inception-v4,
  inception-resnet and the impact of residual connections on learning,'' in
  \emph{Thirty-first AAAI conference on artificial intelligence}, 2017.

\end{thebibliography}

\end{document}